\documentclass[final,5p,times,twocolumn]{elsarticle}
\usepackage{algorithm}
\usepackage{algorithmic}
\usepackage{subcaption} % for 'subtable' env
\usepackage{graphicx}
\usepackage{amsmath}
\usepackage{graphicx}
\usepackage{booktabs}
\usepackage[table]{xcolor}
\usepackage{amssymb}
\usepackage{newfloat}
\usepackage{listings}
\usepackage{wrapfig}
\usepackage{hyperref}
\usepackage{multirow} 
\usepackage[table]{xcolor} 
\usepackage{array}    
\journal{Pattern Recognition}
\begin{document}

\begin{frontmatter}

\title{Physics-inspired Pseudo Anomaly Generation and Prototype Feature Guidance for 3D Anomaly Detection} %% Article title

\author[a,b]{Jian Ning}
\ead{ningjian@whu.edu.cn}
 
\author[a,b]{Qin Zou\corref{cor1}}
\ead{qzou@whu.edu.cn} 

\author[a]{Linchun Wu}
%\ead{linchun.wu@whu.deu.cn} 

\author[c]{Yuanhao Yue}
%\ead{yhyue@ecjtu.edu.cn} 

\author[d]{Kunmo Li}
%\ead{likunmo@mail.dlut.edu.cn} 

\author[b]{Shoubin Chen}
%\ead{shoubin.chen@whu.edu.cn} 

\author[a]{Zhongyuan Wang}
%\ead{zywang@whu.edu.cn} 

\address[a]{School of Computer Science, Wuhan University, Wuhan 430072, Hubei, China}

\address[b]{Guangdong Laboratory of Artificial Intelligence and Digital Economy (SZ), Shenzhen 518060, Guangdong, China}

\address[c]{School of Information and Software Engineering, East China Jiaotong University, Nanchang 330013, Jiangxi, China}

\address[d]{School of Computer Science and Technology, Dalian University of Technology, Dalian 116024, Liaoning, China}

%\tnotetext[1]{This research was supported by the National Natural Science Foundation of China under grant No.~62171324.}

\cortext[cor1]{Corresponding author} 

%% Abstract
\begin{abstract}
3D point cloud anomaly detection plays a vital role in industrial manufacturing, yet it faces significant challenges due to the scarcity and high acquisition cost of real anomalous samples. The inherently anomaly-free training data further hinders detection methods from effectively learning discriminative features between normal and abnormal instances. To address these issues, we propose PA3AD, a novel framework that introduces a physics-inspired pseudo-anomaly generation strategy to create physically plausible anomalous samples from normal data. Additionally, we incorporate prototype features via a weight-sharing mechanism to guide the model in capturing the distribution shifts between normal and anomalous samples. Specifically, PA3AD introduces two key innovations to tackle the scarcity of real anomalies. First, a physics-inspired module generates diverse pseudo-anomalous point clouds from normal data via multi-physics modeling. Second, momentum-updated prototypes and a difference-aware fusion block capture stable normal representations and their discrepancies with pseudo-anomalies. This design effectively learns distribution shifts, achieving superior detection performance. Extensive experiments on the Anomaly-ShapeNet and Real3D-AD datasets demonstrate that our method consistently outperforms existing state-of-the-art approaches. Our code will be made publicly available at \url{https://github.com/NingxiaoJian/PA3AD}.
\end{abstract}


\end{frontmatter}

\section{Introduction}

3D point cloud anomaly detection is widely adopted for automated inspection and quality assurance\cite{wang2025m3dm,xie20243d,yang2024slsg}. Its goal is to localize geometric irregularities from point cloud representations of product surfaces, enabling reliable identification of structural defects in real world inspection. Due to the high cost of anomaly collection and labeling, this task is commonly conducted in an anomaly‑free setting for training. This imbalanced data distribution is a major challenge in 3D point cloud anomaly detection\cite{XIE2024110589,zavrtanik2021draem,bergmann2021mvtec}.

%In recent years, anomaly detection has received increasing attention due to the scarcity and diversity of anomalies in real-world inspection\cite{cao2024complementary}\cite{ZHUANG2025111045}\cite{defard2021padim}. In the 2D image, anomaly detection is often formulated as detecting feature space deviations, with patch-level descriptors and distribution estimation as key ingredients\cite{roth2022towards}\cite{gudovskiy2022cflow}. More recently, vision language priors further extend anomaly detection to open-set categories and cross-scene settings via prompt-based matching\cite{jeong2023winclip}\cite{radford2021learning}. However, extending these 2D paradigms to 3D point clouds is challenging due to the unordered structure, irregular geometry, and density variations. Existing 3D point-cloud methods can be broadly categorized into embedding-based and reconstruction-based paradigms. Embedding-based approaches exploit pretrained features and identify anomalies via feature deviation or retrieval, which is often efficient and sometimes even training-free\cite{wang2023multimodal}\cite{liu2023real3d}\cite{horwitz2023back}. In contrast, reconstruction-based approaches learn a generative or completion model from normal data and localize anomalies using reconstruction residuals\cite{yu2021pointr}\cite{yuan2018pcn}. Beyond these two paradigms, recent work also explores richer spatial cues by explicitly modeling complementary spatial modalities within point clouds, enabling precise detection of both subtle local defects and global deformations\cite{liang2025look}.

Existing 3D point cloud anomaly detection methods\cite{cao2024complementary,ZHUANG2025111045,defard2021padim} are generally divided into two main paradigms: embedding-based and reconstruction-based. Embedding-based approaches\cite{wang2023multimodal,liu2023real3d,horwitz2023back} typically leverage pretrained representations and perform anomaly detection by measuring feature deviations or nearest-neighbor retrieval, which is often efficient and can even be training-free. In contrast, reconstruction-based approaches\cite{yu2021pointr,yuan2018pcn,liang2025look} learn a generative or completion model from normal data and localize anomalies using reconstruction residuals between the input and the reconstructed shape. Some image anomaly detection methods\cite{jeong2023winclip,radford2021learning} have offered new perspectives for point-cloud anomaly detection; however, due to the unordered and sparse nature of point clouds, these methods are difficult to apply directly.

Nevertheless, 3D anomaly detection still faces two main challenges. First, anomalous point clouds are scarce and costly to annotate, which severely limits detection accuracy. In reconstruction-based methods, when training is conducted only on normal samples, anomalous regions tend to be projected back onto the normal manifold, causing anomalies to shrink or even disappear.  Reconstruction-based methods also face the identity shortcut problem\cite{lian2025facing}, in which the network learns to reproduce its input unchanged regardless of whether it is normal or anomalous, making residual-based detection entirely ineffective. Distillation-based methods\cite{lian2026contextual} sidestep both issues by detecting anomalies through teacher-student feature discrepancies, but they lack explicit anomaly supervision for fine-grained localization. A common solution is to introduce synthesized pseudo‑anomalous samples to assist training\cite{li2024towards}. Although recent pseudo-anomaly synthesis methods have shown the value of synthetic supervision, most existing 3D strategies rely mainly on random geometric perturbations or reconstruction residuals, and physics-inspired generation that reflects industrial defect formation remains underexplored for point clouds. However, this also brings new challenges, as their morphology and texture can deviate from real anomalies and may induce bias if not properly handled\cite{zhang2024realnet,schluter2022natural}. Second, perceiving the differences between normal and anomalous features is critical for improving detection accuracy. For example, slight dents can be highly similar in local geometry to normal stamping textures. If the embedding relies mainly on curvature or normal variations, the two may overlap in feature space and cause a substantial drop in detection accuracy\cite{deng2022anomaly}. To the best of our knowledge, these issues have not been addressed by a unified and effective solution, constituting a key bottleneck for bringing point cloud anomaly detection into practical deployment.

In this paper, we propose PA3AD to address the above challenges. It adopts an offset prediction strategy that estimates the displacement vector from each point to the underlying surface, which guides the model to focus on anomalous regions and enhances interpretability. Specifically, we introduce a physics-inspired pseudo-anomaly generation method that synthesizes anomalies according to industrial defect formation mechanisms, thereby addressing the quality issues in pseudo-anomaly generation. We then feed both normal samples and the generated pseudo-anomalies into the detection pipeline. Our weight-sharing network aligns its embeddings in a unified representation space and uses momentum-based normal prototypes as stable anchors. In addition, we embed a local-global attention module in the backbone so that the model can represent global contours and local details. The concatenated features are then passed to a difference-aware fusion module to distinguish between normal and anomalous features, and offset prediction is used to precisely localize anomalous regions.

The contributions of this study are as follows:

$\bullet$ We propose a pseudo‑anomaly generation framework grounded in the physics of industrial defect formation that enforces physical continuity and geometry‑aware deformations, producing pseudo-anomalies with more realistic spatial distributions than geometry‑only methods.

$\bullet$ We develop a momentum-based prototype feature-guided mechanism that preserves stable normal feature prototypes and provides a reliable normality anchor for anomaly detection.

%$\bullet$ We design a local-global attention method that uses local attention with learnable relative positional encoding and an efficient global attention to capture local relations and model global structure.

$\bullet$ We present a difference-aware fusion block and a difference-aware offset loss, thereby converting feature differences into per‑point geometric offsets and enabling robust feature separation and offset localization.

\section{Related work}

\subsection{2D anomaly detection}

With the continuous development of anomaly detection\cite{8517148}, many methods have been proposed to address this problem. Flow-based methods\cite{gudovskiy2022cflow,rudolph2021same} use invertible mappings to maximize normal feature likelihood for pixel-level localization. Teacher-student methods\cite{zhang2023destseg,yamada2022reconstructed} use the deviation between a pretrained teacher and a student as the anomaly score. Memory-bank methods\cite{bae2023pni,kim2023fapm}, exemplified by PatchCore\cite{roth2022towards}, perform nearest-neighbor matching against stored normal features. Reconstruction-based methods\cite{gong2019memorizing,zavrtanik2021reconstruction} take the residual between input and reconstruction as the anomaly score. Synthesis-based approaches\cite{zhang2024realnet,schluter2022natural} insert defects into normal images to construct pseudo-anomalous samples for training. These methods offer transferable insights for 3D anomaly detection; however, the unordered structure and non-uniform density of point clouds make direct application highly challenging.

\subsection{3D anomaly detection}

3D point cloud anomaly detection methods\cite{LIANG2026112924,MiniShift_Simple3D,yu2021pointr} have advanced rapidly with the release of datasets such as Anomaly-ShapeNet\cite{li2024towards} and Real3D-AD\cite{liu2023real3d}. Compared with image anomaly detection, point-cloud-based methods must cope with unordered and irregular structures, sensor noise, and uneven density. BTF\cite{horwitz2023back} directly matches 3D geometric descriptors, establishing a basic framework for 3D anomaly detection. Autoencoding-based methods\cite{cheng2025mc3d} use memory-bank nearest-neighbor metrics to measure feature deviations, but their effectiveness can be limited under non-uniform point densities. AST\cite{rudolph2023asymmetric} amplifies teacher-student output discrepancies at anomalous regions\cite{bergmann2020uninformed}, but its performance depends on likelihood calibration and backbone stability. M3DM\cite{wang2023multimodal} enhances robustness and accuracy by fusing complementary multimodal information, yet it relies heavily on pretrained models and a memory bank. MC3D-AD\cite{cheng2025mc3d} introduces a unified, geometry-aware reconstruction model for class-agnostic reconstruction and cross-category anomaly detection.

Recent studies have explored how to construct anomalies for 3D anomaly detection. IMRNet\cite{li2024towards} synthesizes pseudo anomalies through random local geometric transformations. This shows the value of generated supervision, but such perturbations do not explicitly reflect the continuous surface deformation of real defects. R3D-AD\cite{zhou2024r3d} uses generated local abnormal patches to reduce the synthetic-real gap and then scores anomalies with reconstruction residuals, but these anomalies mainly serve reconstruction rather than physical defect modeling. PO3AD\cite{ye2025po3ad} learns point-wise offsets from pseudo anomalies to normal surfaces, yet its pseudo anomalies are still based on simple geometric perturbations. These studies indicate that pseudo-anomaly construction is an important source of supervision for 3D anomaly detection, but existing strategies still mainly rely on random or simple geometric perturbations, leaving defect-related deformation priors insufficiently modeled.

Although the above methods have made significant progress in 3D anomaly detection, practical deployment remains constrained by two persistent limitations. First, the scarcity of real anomalous point clouds forces anomaly‑free training, which may introduce bias and fail to reflect physically plausible defect distributions. Second, unordered and non‑uniform sampling makes it difficult to capture subtle, nonlinear differences between normal and anomalous geometry. These gaps call for supervisory signals that are physically plausible and continuous, and for feature‑learning mechanisms that adaptively emphasize the most discriminative feature dimensions while remaining robust to scale and sampling.

\section{Method}

\subsection{Problem definition}
Given a training dataset of point clouds containing only normal samples $D_{\text {train}}=\left\{P_{i} \in R^{N \times 3}\right\}_{i=1}^{M}$, where each point cloud $P_{i}$ consists of $N$ 3D points, and $M$ is the number of training samples. The test dataset $D_{\text {test }}=\left\{\left(P_{k}, t_{k}\right)\right\}_{k=1}^{K}$ contains $K$ point clouds with labels $t_{k} \in\{0,1\}$, where 0 denotes normal and 1 denotes anomalous. The goal of point cloud anomaly detection is to learn an anomaly scoring function $\phi: R^{N \times 3}  \rightarrow R^{N}$ which outputs point-level anomaly scores $S=\left\{s_{1}, s_{2}, \ldots, s_{K}\right\}$ for each test point cloud $P_{k}$, where higher scores indicate higher likelihood of being anomalous. 

Additionally, an object-level anomaly score $s_{k}^{(o)}=f\left(S_{k}\right)$ is computed to quantify the overall abnormality of the point cloud. Only normal point clouds $D_{\text {train }}$ are accessible during training, and both $S_{k}$ and $s_{k}^{(o)}$ are required for each unknown test point cloud $P_{k}$ during testing.

\begin{figure*}[!t]
	\centering
	\includegraphics[width=1\textwidth]{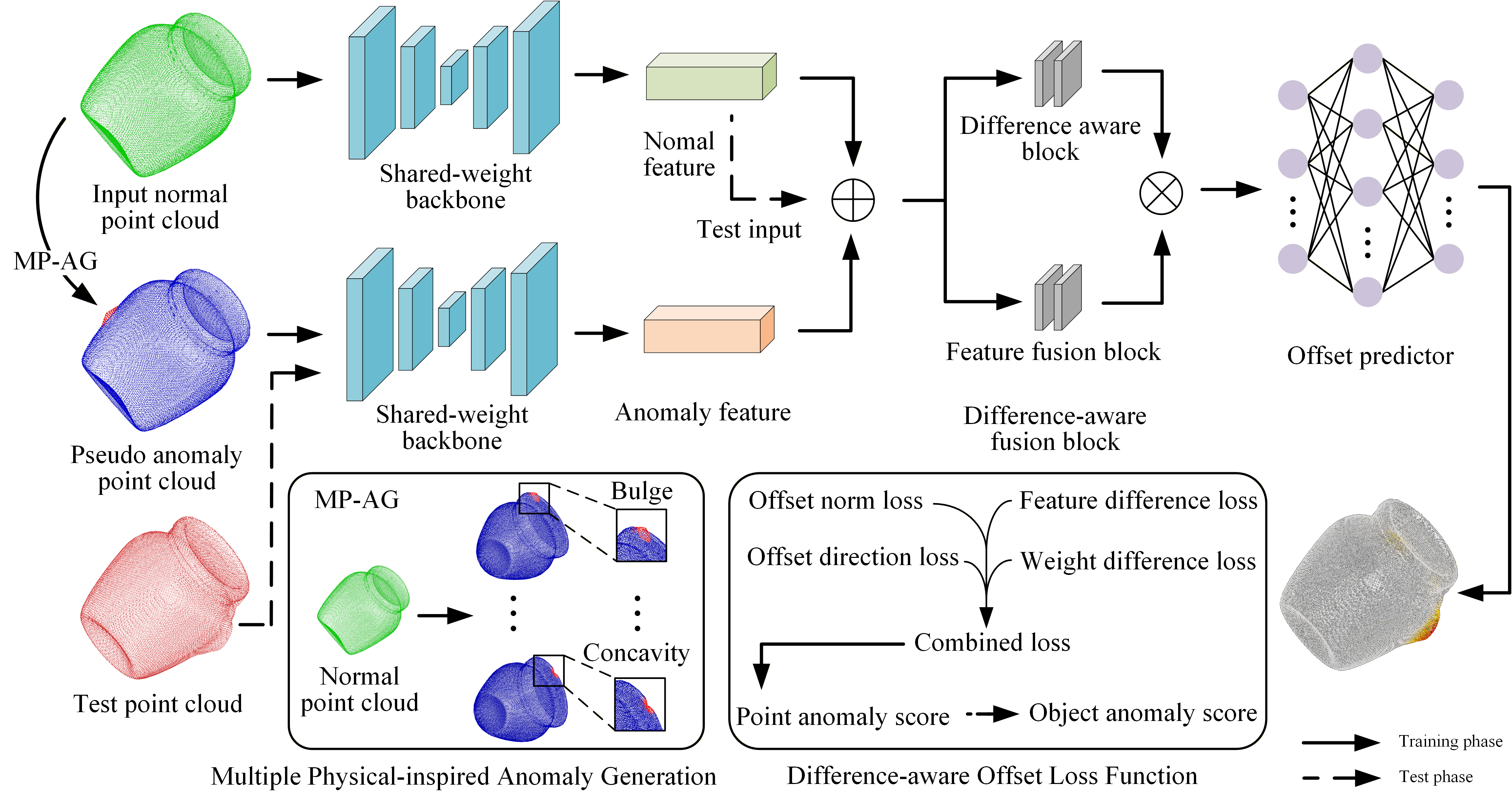}
	\caption{
		The framework of the proposed PA3AD. During training (from upper left to right), a normal point cloud is fed to MP‑AG, which generates a pseudo‑anomalous counterpart for supervision. Normal and anomalous features are extracted by a shared‑weight backbone, concatenated, and passed to the difference‑aware fusion module. An offset predictor is trained to estimate point‑wise displacements between the two clouds. PA3AD is optimized with a difference‑aware offset loss, and an anomaly score is computed. During testing (from lower left to right), the test cloud is passed through the pipeline, combined with the learned normal features, and anomaly predictions are produced.
	}\label{framework}
\end{figure*}

 \subsection{Pipeline overview}

%We propose PA3AD, a framework for anomaly detection in 3D point clouds, as shown in Fig.\ref{framework}. With the aid of a unified pseudo-anomaly generation method based on physics-inspired multi-mechanism modeling, we generate an anomalous point cloud $P_{anomaly}$ for training from a normal point cloud $P_{normal}$.  Both $P_{normal}$ and $P_{anomaly}$ are voxelized and fed into a shared-weight backbone to extract normal and anomalous features, with a local-global attention module incorporated to enhance feature extraction. The features are concatenated and fed into the difference-aware fusion block, which calculates the similarities and differences between features. Then we optimize the entire network with a difference‑aware offset loss. In addition, to reduce the impact of noise on normal features during training,  we introduce a momentum-based normal prototype as a stable anchor and use it as the normal feature input during inference; details are provided in Section \ref{section3.5}.

%Furthermore, the pseudo‑anomaly generation method is elaborated in Section \ref{section3.3}; the shared-weight backbone in Section \ref{section3.4}; the local-global attention block in Section \ref{section3.6}; and the difference‑aware fusion block with the loss function in Section \ref{section3.7}.

We propose PA3AD, a framework for anomaly detection in 3D point clouds, as shown in Fig.\ref{framework}. During training, our physics-inspired MP‑AG module generates a pseudo‑anomalous point cloud $P_{anomaly}$ from a normal point cloud $P_{normal}$. Both point clouds are voxelized and encoded by a shared-weight backbone with a local-global attention module. The resulting normal and anomalous features are concatenated, fused by the difference-aware block, and supervised via a difference‑aware offset loss for offset prediction. During inference, we use a momentum-updated normal prototype as a stable normal anchor. The following subsections describe each component in detail.

\subsection{Multiple physics-inspired pseudo‑anomaly generation}
%\subsection{Multiple physical-based pseudo‑anomaly generation}
\label{section3.3}
%In order to improve the generalization of point cloud anomaly detection models to various anomaly morphologies, we propose a multiple physical-based pseudo-anomaly generation method called MP-AG. It creates physics‑constrained pseudo-anomalies on normal point clouds, increasing distributional diversity and yielding more informative gradients for optimization.

From an industrial perspective, we focus on five morphological primitives: bulge, concavity, hole, crack, and bend. These correspond to high-frequency defect families in discrete manufacturing such as sheet-metal stamping, casting, and injection molding, and are widely represented in public 3D inspection benchmarks\cite{li2024towards,liu2023real3d}. They span local outward and inward displacement, material removal, in-plane structural discontinuities, and low-frequency shape drift, providing a compact basis for geometric deformation modeling under real-world conditions.

Given an input normal point cloud sample $P=\left\{p_{i} \in R^{3}\right\}_{i=1}^{N}$, where $N$ denotes the number of points in the point cloud, and $p_{i}$ denotes the 3D coordinates of the $i‑th$ point. Since the material properties and boundary conditions required for rigorous physical simulation are unavailable in anomaly detection settings, we instead design parameterized deformation functions that preserve the key morphological signatures of each defect type. Bulge and concavity are among the most common anomaly types, typically manifesting as local protrusions or depressions. Because such defects exhibit a peak at the center with smooth radial decay, we adopt the Gaussian RBF for its inherent localized and symmetric attenuation. In order to generate anomalous points $p_{i}^{\prime}$, we employ this radial basis function along the surface-normal direction of the selected region:
\begin{equation}
p_{i}^{\prime}=p_{i}+\alpha \cdot \exp \left(-\frac{\left\|p_{i}-c\right\|^{2}}{2 \sigma_{b h}^{2}}\right) \cdot n_{i},
\end{equation}
where $\alpha$ denotes the perturbation magnitude, with positive values indicating bulges and negative values indicating concavities. $\sigma_{bh}$ controls the spatial extent of influence, $c$ denotes the center of the selected deformation region, and $n_i$ denotes the surface normal vector. %Unlike geometric perturbations, our RBF‑based formulation enforces continuous local deformation, yielding a smoother transition between anomalous and normal regions.

Hole‑type anomalies commonly arise from material loss or perforation, where the center region is completely missing and the surrounding points sink gradually toward the boundary. Since $tanh$ saturates in the center and provides a smooth yet compact transition, we adopt it to model this spatial pattern:
\begin{equation}
p_{i}^{\prime}=p_{i}-\beta \cdot \tanh \left(\frac{\left\|p_{i}-c\right\|}{\sigma_{h}}\right) \cdot n_{i},
\end{equation}
where $\beta$ denotes the sinking depth, $\sigma_{h}$ is a smoothing coefficient, and $c$ denotes the hole center. We remove points satisfying $\left\|p_{i}-c\right\|<r_{h}$, where $r_{h}$ denotes the hole radius.

Crack‑type anomalies usually manifest as slender linear fractures. According to fracture mechanics, cracks propagate along the principal stress direction, but computing true stress fields requires material parameters unavailable in our setting. We approximate this direction via PCA on the $k$-nearest neighbors of the crack center, taking the largest-eigenvalue eigenvector as ${v}$, which corresponds to the most elongated local direction where the structure is thinnest and weakest perpendicular to it. We then slice the point cloud linearly along this direction with a slight interfacial offset:
\begin{equation}
p_{i}^{\prime}=\left\{\begin{array}{l}
p_{i}+\gamma \cdot {v}, \text { if }\left(p_{i}-c\right)^{T} {v}>0, \\
p_{i}-\gamma \cdot {v}, \text { otherwise, }
\end{array}\right.
\end{equation}
where $\gamma$ denotes the crack width, and $c$ denotes the crack center.

Bending‑type anomalies generally manifest as smooth, axis-aligned curvature changes across a local region. Inspired by the deflection profile described by Euler-Bernoulli beam theory, we design a deformation function that produces continuous axial displacement, sampling the deflection at discrete point locations rather than solving the underlying PDE:
\begin{equation}
p_{i}^{\prime}=p_{i}+\eta \cdot f_{\text {bend }}\left(p_{i}, c, a\right),
\end{equation}
\begin{equation}
f_{b e n d}\left(p_{i}, a\right)=w_{i} \frac{a \times d_{i, \perp}}{\left\|d_{i, \perp}\right\|+\epsilon},
\end{equation}
where $\eta$ denotes the bending strength, $a$ denotes the bending axis (unit direction), $\epsilon$ is a small constant for numerical stability, $f_{b e n d}$ denotes the bending deformation function, $d_{i, \perp}$ denotes the vector from $p_{i}$ to the bending axis, and $w_{i}$ denotes the smoothing weight. We compute the displacement direction of point $p_{i}$ in the normal plane to the bending-axis using $a \times d_{i, \perp}$, and the displacement magnitude is determined by the bending strength and distance‑based weighting, thereby ensuring spatial continuity and physical plausibility.

%Because MP‑AG is inspired by industrial defect formation mechanisms rather than unconstrained geometric perturbations, the synthesized anomalies  preserve smoother geometric transitions at defect-normal interfaces and capture defect-relevant curvature and normal patterns. This provides supervision that is closer to the statistics of real defects and thus improves transfer to real test sets. Concretely, bulge and concavity are realized by smooth, normal-aligned RBF deformations, hole and crack capture material removal, and bend follows the Euler-Bernoulli kinematic field. These primitives align with the dominant defect categories commonly encountered in real-world industrial inspection, and our scale‑aware parameterization matches the empirical ranges of defect height, radius and width across categories. As a result, models trained with MP‑AG generalize more reliably to real anomalies than those trained with geometry-only perturbations. Our generated anomalous samples will be presented in detail and experimentally validated in Section \ref{section4.6}.

Because each deformation function in MP-AG is shaped by the morphological characteristics of a specific defect type, the synthesized anomalies exhibit more structured and physically plausible geometry than those produced by unconstrained random perturbations.

\subsection{Shared-weight backbone feature extraction}
\label{section3.4}
\begin{figure*}
	\centering
	\includegraphics[width=0.98\textwidth]{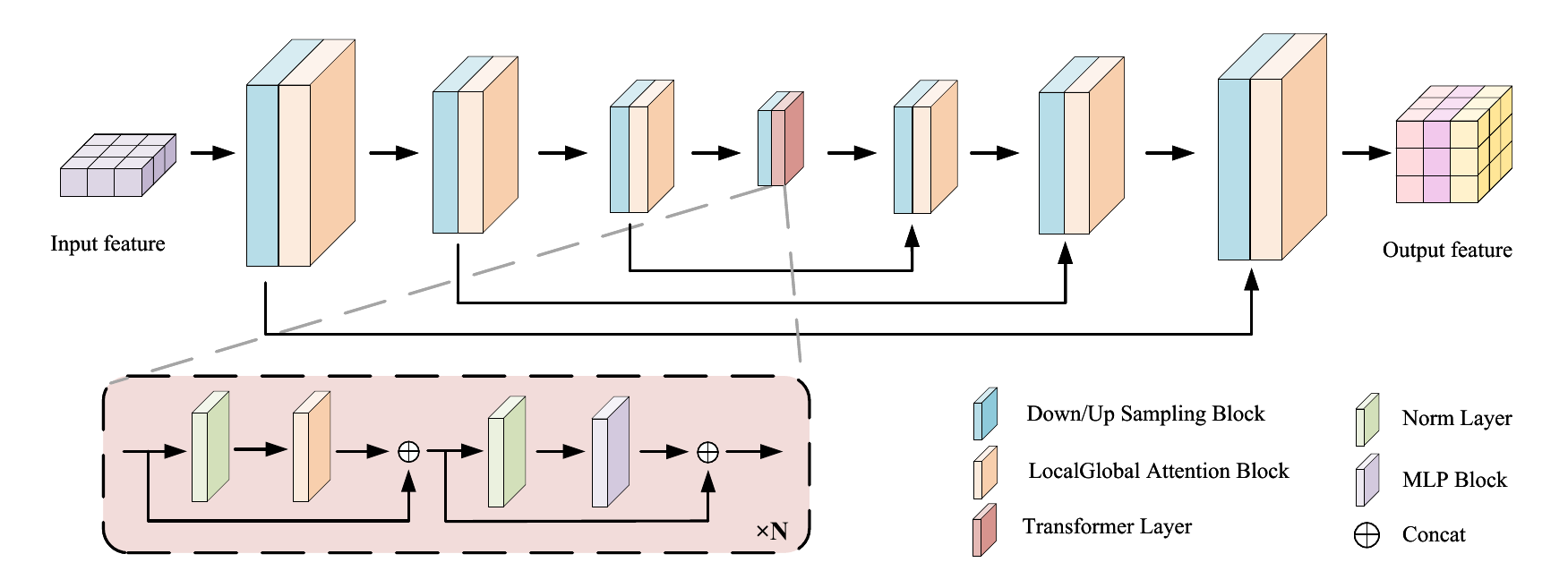}
	\caption{
		The pipeline of our shared-weight backbone. We take voxelized point cloud features as input and use an encoder-decoder to progressively extract semantics from local to global. Each block combines sparse convolutions with a local-global attention module. At the bottleneck layer, we stack two transformer layers with local-global attention, as shown in the lower left.
	}
	\label{tab:shared-weight backbone}
\end{figure*}

%The key to anomaly detection is the ability to extract local geometric features and spatial structure efficiently and robustly. Therefore, we propose a shared-weight feature extraction backbone that encodes normal and pseudo-anomalous point clouds with shared parameters, which ensures a consistent feature space and improves parameter efficiency, thereby improving both training and inference efficiency. Our network architecture is shown in Fig.\ref{tab:shared-weight backbone}. 
We propose a shared-weight feature extraction backbone that encodes normal and pseudo-anomalous point clouds with shared parameters, ensuring a consistent feature space and improving both training and inference efficiency. Our network architecture is shown in Fig.\ref{tab:shared-weight backbone}.

For the input point cloud $P=\left\{p_{i} \in R^{3}\right\}_{i=1}^{N}$, we first apply a voxelization function $V$ to convert it into a sparse tensor representation:
\begin{equation}
X=V(P, s)=(C, {G}),
\end{equation}
where $s$ denotes the voxel size, $C$ denotes the coordinates of non‑empty voxels, and ${G}$ denotes the corresponding aggregated features.

Since conventional convolutions struggle to capture long-range geometric relations, we introduce two transformer blocks with local-global attention at the bottleneck layer, as highlighted in red at the bottom left of Fig.\ref{tab:shared-weight backbone}. Formally, this can be expressed as:
\begin{equation}
Z_{i}^{a t t}=Z_{0}+\operatorname{Attn}\left(\operatorname{LN}\left(Z_{0}\right)\right),
\end{equation}
\begin{equation}
Z_{i+1}=Z_{i}^{a t t}+M L P\left(L N\left(Z_{i}^{a t t}\right)\right),
\end{equation}
where $Z_{0}$ denotes the input features, LN denotes the LayerNorm layer, and Attn denotes our local-global attention block, which is elaborated in Section \ref{section3.6}.

\subsection{Momentum-based normal feature generation}
\label{section3.5}
We are inspired by the human brain’s way of performing anomaly detection. For example, when judging whether an apple is abnormal, people typically first form a mental prototype of a normal apple and then compare the apple in hand against this prototype. Importantly, normality is not a static concept. It is progressively shaped and updated by continuous exposure, yielding a stable yet adaptive reference against which anomalies are identified. In representation learning terms, this suggests maintaining a prototype embedding as a stable anchor of normality. Furthermore, in order to preserve long-term normal statistics while suppressing noise and non-stationary fluctuations, we introduce a momentum‑based normal prototype as a stable reference for normal features.

Specifically, we gradually increase the momentum coefficient from 0.1 to 0.999, which corresponds to a gradual decrease in the update rate of features during training. This means that at the beginning of training, the prototype rapidly adapts to the normal features, whereas in later stages it changes slowly and thus serves as a stable anchor. Below we provide a brief analysis to clarify the behavior of this exponential moving average (EMA) update under mild assumptions.

In terms of dynamics, our prototype update can be written as an EMA recursion with initial condition ${z}^{(0)}=0$:
\begin{equation}
	{z}^{(t+1)}=\mu(t) {z}^{(t)}+(1-\mu(t)) f^{(t)},
\end{equation}
where ${z}^{(t)} \in \mathbb{R}^{d}$ denotes the prototype after the $t$‑th update, $f^{(t)} \in \mathbb{R}^{d}$ denotes the batch mean normal feature at iteration $t$, and $\mu(t) \in[0,1)$ is the momentum coefficient.

Unrolling the recursion shows that ${z}^{(t)}$ is a weighted average of past features:
\begin{equation}
	{z}^{(t)}=\sum_{\tau=0}^{t-1}\left((1-\mu(\tau))\prod_{j=\tau+1}^{t-1}\mu(j)\right) f^{(\tau)}.
\end{equation}
Hence, when $\mu(t)$ increases during training, older features receive larger aggregate weight and the prototype becomes progressively smoother.

To discuss convergence, we consider a standard setting where the batch feature can be decomposed as
\begin{equation}
	f^{(t)}=\mu_{\text{gt}}+\xi^{(t)},
\end{equation}
where $\mu_{\text{gt}} \in \mathbb{R}^{d}$ is the (asymptotic) mean normal feature and $\xi^{(t)}$ is a zero‑mean perturbation capturing minibatch noise and non‑stationarity. Define the residual
\begin{equation}
	\varepsilon^{(t)}={z}^{(t)}-\mu_{\text{gt}}.
\end{equation}
Then the residual evolves as
\begin{equation}
	\varepsilon^{(t+1)}=\mu(t)\varepsilon^{(t)}+(1-\mu(t))\xi^{(t)}.
\end{equation}
If $\mu(t)\le \bar{\mu}<1$ and the perturbation has bounded second moment, then $\varepsilon^{(t)}$ remains bounded and the EMA acts as a low‑pass filter whose effective averaging window is approximately $1/(1-\mu(t))$. In particular, when $\mu(t)$ becomes close to 1 in later training, the update becomes conservative and suppresses minibatch noise, yielding a stable prototype that improves robustness at inference time.

The normal prototype is essentially a feature vector that is continuously updated via EMA during training. After training is completed, the converged prototype feature is saved as part of the model. During inference, the saved prototype feature is directly loaded and used as the normal reference to compare against the test sample features, which are then fed into the difference-aware fusion block to produce offset predictions and anomaly scores.

\begin{figure*}
	\centering
	\includegraphics[width=1\textwidth]{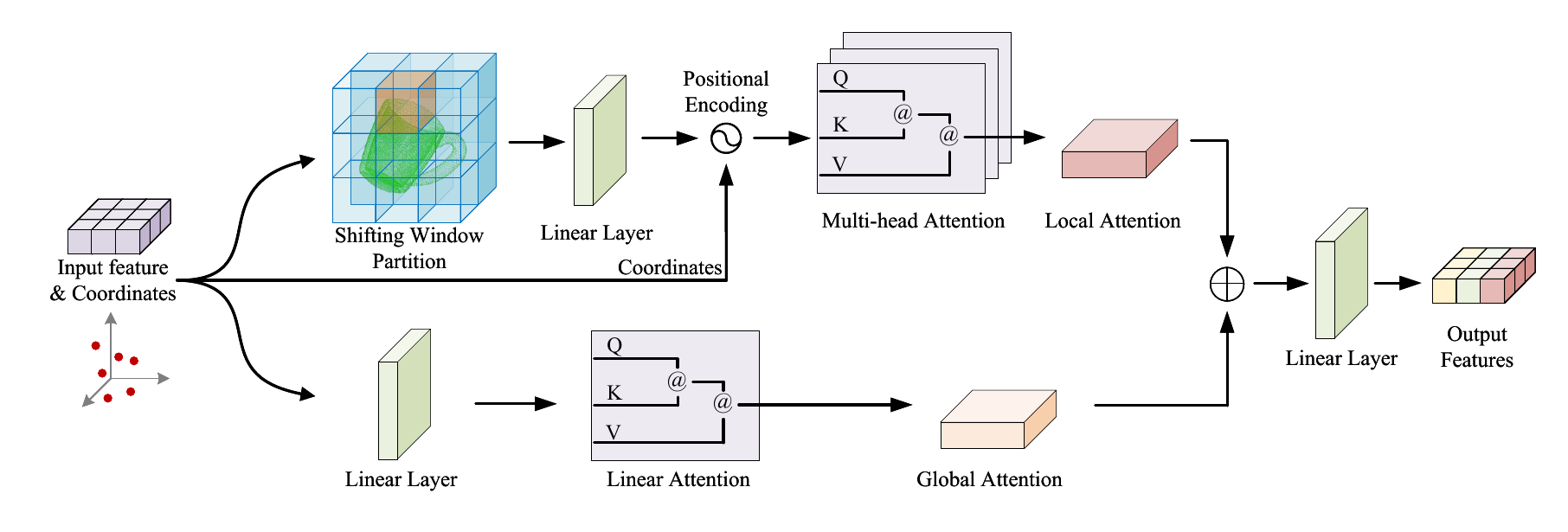}
	\caption{
    The implementation of local-global attention. In the local branch, we partition the point cloud using a shifted sliding window, then apply linear projections, positional encoding, and multi-head self-attention to produce local features. To control the complexity of the global branch, we use linear projections with linear attention. We finally fuse the two branches to generate the output features.
	}
	\label{tab:local-global attention}
\end{figure*}

\subsection{Local-global attention block}
\label{section3.6}
 %To overcome the limited receptive field of conventional 3D convolutions, we propose a local-global attention method. By processing local geometric details and global structural context in parallel, it enables comprehensive and efficient modeling of point cloud features. Local attention computes self‑attention within spatially partitioned local windows and, with positional encoding, precisely captures positional relationships between voxels. Global attention models relationships among all voxels across the entire point‑cloud space, thereby capturing the object’s overall shape, symmetry, and structural integrity. Our network architecture is shown in Fig.\ref{tab:local-global attention}.
 To overcome the limited receptive field of conventional 3D convolutions, we propose a local-global attention method that models local geometric details and global structural context in parallel. Our network architecture is shown in Fig.\ref{tab:local-global attention}.

Our local attention block partitions the input feature matrix $Z \in R^{M \times d}$ into $K$ cubic windows with side length $w$ and stride $w/2$, forming subsets $W_{k} \subset Z$ of size $M_{k} \times d$ ($M_{k} \leq w^{3}$). To recover spatial positional relationships among points, we propose a learnable positional encoding. For the point‑cloud coordinates $P^{(i)} \in R^{w \times 3}$ within the $i‑th$ window, we have:
\begin{equation}
R_{j, k}^{(i)}=P_{j}^{(i)}-P_{k}^{(i)},
\end{equation}
where $P_{j}^{(i)}$ and $P_{k}^{(i)}$ denote the 3D coordinates of the $j‑th$ and $k‑th$ points within the window, respectively, and $R_{j, k}^{(i)}$ denotes the relative coordinates between points. We then feed the resulting relative position matrix $R^{(i)}$ into the positional encoding matrix $F_{p o s}$ to obtain the positional features $F_{p o s}^{(i)}$. Our positional encoding network uses two MLP layers:
\begin{equation}
F_{p o s}^{(i)}=F_{p o s}\left(R^{(i)}\right).
\end{equation}
Then we compute multi‑head self‑attention within each window $W_{k}$. For the $h‑th$ attention head:
\begin{equation}
\text { head }_{k, h}=\text {softmax}\left(\frac{\left(W_{k} W_{h}^{Q}\right)\left(W_{k} W_{h}^{K}\right)^{T}}{\sqrt{d_{k}}}+F_{\text {pos }}^{(k)}\right)\left(W_{k} W_{h}^{V}\right),
\end{equation}
where $W_{h}^{Q}$, $W_{h}^{K}$, and $W_{h}^{V} \in R^{d \times d_{k}}$ are learnable projection matrices, and the local‑attention output $Z_{\text {local }} \in R^{M \times d}$ is obtained by concatenating the outputs of all heads.

Our global attention adopts linear attention to capture global context with $\mathrm{O}(N)$ complexity. Given an input feature matrix $Z \in R^{M \times d}$, we first apply linear transformations to obtain the corresponding matrices:
\begin{equation}
\left\{\begin{array}{l}
Q=Z W_{Q}, \\
K=Z W_{K}, \\
V=Z W_{V}.
\end{array}\right.
\end{equation}
Consequently, we define the global attention $Z_{\text {global }}$ as:
\begin{equation}
Z_{\text {global }}=D^{-1}\left(\phi(Q) \phi(K)^{T} V\right),
\end{equation}
where $W_{Q}$, $W_{K}$, and $W_{V} \in R^{d \times d}$ are learnable projection matrices. We use ReLU as the feature‑mapping function $\phi$, and $D$ denotes a normalized diagonal matrix.

\subsection{Difference‑aware fusion block and loss function}
\label{section3.7}
After extracting normal $F_{\text {normal }} \in R^{N \times d}$ and anomalous $F_{\text {anomaly }} \in R^{N \times d}$ features from the shared-weight backbone, we introduce a difference-aware fusion block that learns adaptive weights to amplify informative differences. The difference‑enhanced features are translated by an offset head into pointwise geometric displacements, supervised by a difference-aware offset loss.

We first concatenate $F_{\text{normal}}$ and $F_{\text{anomaly}}$ to obtain $F_{\text{concat}} \in \mathbb{R}^{N \times 2d}$, which is fed into the feature fusion block. Meanwhile, we compute the raw feature difference as $\Delta F = F_{\text{anomaly}} - F_{\text{normal}}$ and use $\Delta F$ as the input to the difference-aware branch to produce adaptive weights that highlight key discrepancies. Our feature fusion block comprises two fully connected layers that compress the feature dimensionality from $2d$ to $d$, enabling the model to learn fused features and output a $d$-dimensional fused representation $F_{\text {fuse }}$: 
\begin{equation}
	F_{\text {fuse }}=\operatorname{Linear}\left(\operatorname{RELU}\left(\operatorname{Linear}\left(F_{\text {concat }}\right)\right)\right).
\end{equation}

Inspired by attention mechanisms, our difference‑aware module generates difference-aware adaptive weights $w_{\text {diff }}$ via fully connected layers and multiplies them with the fused features to obtain the difference‑enhanced features $F_{\text {diff}}^{\text {enh}}$:
\begin{equation}
	w_{\text {diff }}=\operatorname{Sigmoid}\left(\operatorname{Linear}\left(\operatorname{RELU}\left(\operatorname{Linear}\left(\Delta F\right)\right)\right)\right),
\end{equation}
\begin{equation}
	F_{\text {diff}}^{\text {enh}}=F_{\text {fuse}} \otimes w_{\text{diff}}.
\end{equation}
Then we estimate point-wise displacements from the point cloud using offset prediction. We use MLP to learn the mapping from feature differences to geometric space. Through linear transformation and PReLU activation, we extract key anomalous pattern information, progressively convert high-dimensional difference features into low-dimensional geometric offset representations, and finally output the offset prediction $\hat{o} \in R^{3}$. 

After obtaining the difference features $F_{\text {diff}}^{\text {enh}}$ and the offset predictions $\hat{o}$, we supervise the entire network with our difference‑aware offset loss function. First, to ensure that the magnitude of the predicted offset vectors better matches the synthesized anomalies, we use L1 and L2 as offset-norm loss:
\begin{equation}
L_{\text {norm }}=\frac{1}{N} \sum_{i=1}^{N}\left(\left\|\hat{o}_{i}-o_{i}\right\|_{1}+\left\|\hat{o}_{i}-o_{i}\right\|_{2}\right),
\end{equation}
where $\hat{o}_{i} \in R^{3}$ denotes the offset prediction for the $i-th$ point, ${o}_{i} \in R^{3}$ denotes the offset of the generated point cloud, and for normal points, $o_{i}=0$.

Besides considering the offset magnitude, we also consider the offset direction:
\begin{equation}
L_{d i r}=\frac{1}{N} \sum_{i=1}^{N}\left(1-\frac{\hat{o}_{i} \cdot o_{i}}{\left\|\hat{o}_{i}\right\|_{2}\left\|o_{i}\right\|_{2}+\varepsilon}\right),
\end{equation}
where $\varepsilon=1 \times 10^{-8}$ is a constant to prevent division by zero. When the predicted offset direction aligns with the generated anomaly direction, $L_{d i r}=0$.

Since anomalies constitute only a small part of the point cloud, to prevent normal and anomalous features from becoming overly similar during training, we introduce a feature‑discrepancy regularization loss:
\begin{equation}
L_{\text {feat }}=\exp \left(-\frac{1}{N} \sum_{i=1}^{N}\left\|F_{n}^{(i)}-F_{a}^{(i)}\right\|_{2}^{2}\right),
\end{equation}
where $F_{n}^{(i)}$ and $F_{a}^{(i)}$ denote the normal and anomalous features of the $i‑th$ point; whenever $\left\|F_{n}^{(i)}-F_{a}^{(i)}\right\|_{2}^{2} \rightarrow 0$ is satisfied, $L_{\text {feat }} \rightarrow 1$, so that similar features are punished.

In addition, we expect the weights $w_{\text {diff }}$ to reflect differences without completely suppressing feature expression; thus, we propose a difference‑weight regularization loss:
\begin{equation}
L_{\text {weight }}=\frac{1}{N \cdot d} \sum_{i=1}^{N} \sum_{j=1}^{d}\left|w_{\text {diff }}^{(i, j)}-\alpha\right|,
\end{equation}
where $N$ denotes the number of voxels in the current batch, $d$ denotes the feature dimension and $\alpha$ denotes the central value of the weights. To prevent weight skew, we set $\alpha = 0.5$.

Finally, our difference‑aware offset loss function is:
\begin{equation}
L=L_{\text {norm }}+L_{\text {dir }}+\lambda_{1} L_{\text {feat }}+\lambda_{2} L_{\text {weight }},
\end{equation}
where $\lambda_{1}=0.01$ and $\lambda_{2}=0.001$.
The point‑level anomaly score $s\left(p_{i}\right)$ is defined as:
\begin{equation}
s\left(p_{i}\right)=\left|\hat{o}_{i, x}\right|+\left|\hat{o}_{i, y}\right|+\left|\hat{o}_{i, z}\right|.
\end{equation}
The object‑level anomaly score is defined as:
\begin{equation}
s^{(o)}(P)=\frac{1}{N} \sum_{i=1}^{N} s\left(p_{i}\right).
\end{equation}
The higher the score, the more likely the sample is anomalous. The total loss $L$ jointly supervises offset regression and feature separation, enabling precise point-level anomaly scoring.

\section{Experiments}

\subsection{Datasets}
We conducted experiments on Anomaly‑ShapeNet~\cite{li2024towards} and Real3D-AD~\cite{liu2023real3d}. \\
Anomaly-ShapeNet is a synthetic dataset covering 40 categories with 1,600 samples; each category provides only four normal training samples and a mixture of normal and defective test samples. Real3D‑AD comprises high‑resolution point clouds of real industrial objects across 12 categories, each with four normal training samples and 100 test instances. Notably, Real3D‑AD exhibits a pronounced acquisition domain gap: training samples are captured via 360° turntable scanning, whereas test samples are predominantly single‑sided scans.

\subsection{Experimental Settings}

During training, for each input normal sample, we generate pseudo‑anomalous samples using the MP‑AG method proposed in Section \ref{section3.3}. We adopt weighted random sampling to select the pseudo-anomaly type for each training sample. The sampling probabilities are empirically set as: Bulge 0.35, Concavity 0.35, Hole 0.10, Crack 0.10, Bend 0.10, with higher weights assigned to protrusion and depression defects as they are more prevalent and challenging to detect. The voxel size is set to 0.05. Training uses the AdamW optimizer with a batch size of 32 and a learning rate of $1\times10^{-3}$ for 2000 epochs. The model contains 39.45M parameters and is trained from scratch without pretraining. Our implementation uses PyTorch 2.3.1 and CUDA 11.8, running on an Intel Core Ultra 9 285K with 64GB RAM and a single NVIDIA RTX 4090. Training takes approximately 12 hours and inference takes 18.2ms per sample.

\begin{table*}[!ht]
	\centering
	\normalsize
	\setlength{\tabcolsep}{2.0pt}
	\renewcommand{\arraystretch}{0.70}
	\caption{O-AUROC performance of different methods on Anomaly-ShapeNet across 40 categories, where best and second-place results are highlighted in \textcolor{red}{\textbf{red}} and \textcolor[rgb]{ 0,  .555,  .976}{\textbf{blue}}, respectively.}
	\label{o-auroc-anomalyshapenet}
	\resizebox{1\textwidth}{!}{
		\begin{tabular}{l|ccccccccccc}
			\toprule
			\multicolumn{12}{c}{\textbf{O-AUROC}} \\
			\midrule
			\textbf{Method} & \textbf{ashtray0} & \textbf{bag0} & \textbf{bottle0} & \textbf{bottle1} & \textbf{bottle3} & \textbf{bowl0} & \textbf{bowl1} & \textbf{bowl2} & \textbf{bowl3} & \textbf{bowl4} & \textbf{bowl5} \\
			\midrule
			\textbf{BTF(Raw) (CVPR23’)} & 0.578 & 0.410 & 0.597 & 0.510 & 0.568 & 0.564 & 0.264 & 0.525 & 0.385 & 0.664 & 0.417 \\
			\textbf{BTF(FPFH) (CVPR23’)} & 0.420 & 0.546 & 0.344 & 0.546 & 0.322 & 0.509 & 0.668 & 0.510 & 0.490 & 0.609 & 0.699 \\
			\textbf{M3DM (CVPR23’)} & 0.577 & 0.537 & 0.574 & 0.637 & 0.541 & 0.634 & 0.663 & 0.684 & 0.617 & 0.464 & 0.409 \\
			\textbf{PatchCore(FPFH) (CVPR22’)} & 0.587 & 0.571 & 0.604 & 0.667 & 0.572 & 0.504 & 0.639 & 0.615 & 0.537 & 0.494 & 0.558 \\
			\textbf{PatchCore(PointMAE) (CVPR22')} & 0.591 & 0.601 & 0.513 & 0.601 & 0.650 & 0.523 & 0.629 & 0.458 & 0.579 & 0.501 & 0.593 \\
			\textbf{CPMF (PR24’)} & 0.353 & 0.643 & 0.520 & 0.482 & 0.405 & 0.783 & 0.639 & 0.625 & 0.658 & 0.683 & 0.685 \\
			\textbf{Reg3D-AD (NeurIPS23’)} & 0.597 & 0.706 & 0.486 & 0.695 & 0.525 & 0.671 & 0.525 & 0.490 & 0.348 & 0.663 & 0.593 \\
			\textbf{IMRNet (CVPR24’)} & 0.671 & 0.660 & 0.552 & 0.700 & 0.640 & 0.681 & 0.702 & 0.685 & 0.599 & 0.676 & 0.710 \\
			\textbf{R3D-AD (ECCV24’)} & 0.833 & 0.720 & 0.733 & 0.737 & 0.781 & 0.819 & 0.778 & 0.741 & 0.767 & 0.744 & 0.656 \\
			\textbf{PO3AD(CVPR25')} & \textcolor[rgb]{1,0,0}{\textbf{1.000}} & 0.833 & 0.900 & 0.933 & 0.926 & 0.922 & 0.829 & \textcolor[rgb]{0,.555,.976}{\textbf{0.833}} & 0.881 & \textcolor[rgb]{0,.555,.976}{\textbf{0.981}} & 0.849 \\
			\textbf{MC3D-AD (IJCAI25’)} & 0.962 & 0.805 & 0.795 & 0.709 & 0.756 & 0.930 & \textcolor{red}{\textbf{0.978}} & 0.719 & 0.885 & 0.911 & 0.754 \\
			\textbf{Reg2Inv(NeurIPS25)} & 0.900 & \textcolor{red}{\textbf{1.000}} & \textcolor{red}{\textbf{1.000}} & \textcolor{red}{\textbf{1.000}} & \textcolor{red}{\textbf{1.000}} & \textcolor{red}{\textbf{1.000}} & 0.807 & 0.656 & 0.585 & 0.852 & 0.818 \\
			\textbf{Simple3D (AAAI26’)} & \textcolor[rgb]{0,.555,.976}{\textbf{0.995}} & 0.881 & 0.976 & \textcolor[rgb]{0,.555,.976}{\textbf{0.951}} & \textcolor{red}{\textbf{1.000}} & \textcolor{red}{\textbf{1.000}} & 0.830 & 0.711 & \textcolor{red}{\textbf{0.911}} & 0.730 & \textcolor[rgb]{0,.555,.976}{\textbf{0.863}} \\
			\rowcolor{gray!10}\textbf{PA3AD(Ours)} & \textcolor{red}{\textbf{1.000}} & \textcolor[rgb]{0,.555,.976}{\textbf{0.938}} & \textcolor[rgb]{0,.555,.976}{\textbf{0.981}} & 0.933 & \textcolor[rgb]{0,.555,.976}{\textbf{0.971}} & \textcolor[rgb]{0,.555,.976}{\textbf{0.989}} & \textcolor[rgb]{0,.555,.976}{\textbf{0.896}} & \textcolor{red}{\textbf{0.956}} & \textcolor[rgb]{0,.555,.976}{\textbf{0.904}} & \textcolor{red}{\textbf{0.989}} & \textcolor{red}{\textbf{0.926}} \\
			\midrule
			
			\midrule
			\textbf{Method} & \textbf{bucket0} & \textbf{bucket1} & \textbf{cap0} & \textbf{cap3} & \textbf{cap4} & \textbf{cap5} & \textbf{cup0} & \textbf{cup1} & \textbf{eraser0} & \textbf{headset0} & \textbf{headset1} \\
			\midrule
			\textbf{BTF(Raw) (CVPR23’)} & 0.617 & 0.321 & 0.668 & 0.527 & 0.468 & 0.373 & 0.403 & 0.521 & 0.525 & 0.378 & 0.515 \\
			\textbf{BTF(FPFH) (CVPR23’)} & 0.401 & 0.633 & 0.618 & 0.522 & 0.520 & 0.586 & 0.586 & 0.610 & 0.719 & 0.520 & 0.490 \\
			\textbf{M3DM (CVPR23’)} & 0.309 & 0.501 & 0.557 & 0.423 & 0.777 & 0.639 & 0.539 & 0.556 & 0.627 & 0.577 & 0.617 \\
			\textbf{PatchCore(FPFH) (CVPR22’)} & 0.469 & 0.551 & 0.580 & 0.453 & 0.757 & 0.790 & 0.600 & 0.586 & 0.657 & 0.583 & 0.637 \\
			\textbf{PatchCore(PMAE) (CVPR22')} & 0.593 & 0.561 & 0.589 & 0.476 & 0.727 & 0.538 & 0.610 & 0.556 & 0.677 & 0.591 & 0.627 \\
			\textbf{CPMF (PR24’)} & 0.482 & 0.601 & 0.601 & 0.551 & 0.553 & 0.697 & 0.497 & 0.499 & 0.689 & 0.643 & 0.458 \\
			\textbf{Reg3D-AD (NeurIPS23’)} & 0.610 & 0.752 & 0.693 & 0.725 & 0.643 & 0.467 & 0.510 & 0.538 & 0.343 & 0.537 & 0.610 \\
			\textbf{IMRNet (CVPR24’)} & 0.580 & 0.771 & 0.737 & 0.775 & 0.652 & 0.652 & 0.643 & 0.757 & 0.548 & 0.720 & 0.676 \\
			\textbf{R3D-AD (ECCV24’)} & 0.683 & 0.756 & 0.822 & 0.730 & 0.681 & 0.670 & 0.776 & 0.757 & 0.890 & 0.738 & 0.795 \\
			\textbf{PO3AD(CVPR25')} & 0.853 & 0.787 & \textcolor[rgb]{0,.555,.976}{\textbf{0.877}} & 0.859 & 0.792 & 0.670 & \textcolor[rgb]{0,.555,.976}{\textbf{0.871}} & 0.833 & \textcolor[rgb]{0,.555,.976}{\textbf{0.995}} & 0.808 & 0.923 \\
			\textbf{MC3D-AD (IJCAI25’)} & 0.898 & 0.784 & 0.793 & 0.701 & 0.835 & 0.761 & 0.743 & \textcolor[rgb]{0,.555,.976}{\textbf{0.952}} & 0.776 & 0.862 & 0.886 \\
			\textbf{Reg2Inv(NeurIPS25)} & 0.813 & \textcolor{red}{\textbf{0.902}} & 0.659 & 0.863 & 0.681 & \textcolor[rgb]{0,.555,.976}{\textbf{0.902}} & 0.733 & 0.933 & \textcolor{red}{\textbf{1.000}} & \textcolor{red}{\textbf{1.000}} & 0.843 \\
			\textbf{Simple3D (AAAI26’)} & \textcolor[rgb]{0,.555,.976}{\textbf{0.959}} & 0.790 & 0.852 & \textcolor[rgb]{0,.555,.976}{\textbf{0.867}} & \textcolor{red}{\textbf{0.912}} & 0.804 & \textcolor{red}{\textbf{1.000}} & 0.824 & \textcolor{red}{\textbf{1.000}} & \textcolor[rgb]{0,.555,.976}{\textbf{0.982}} & \textcolor[rgb]{0,.555,.976}{\textbf{0.957}} \\
			\rowcolor{gray!10}\textbf{PA3AD(Ours)} & \textcolor{red}{\textbf{0.968}} & \textcolor[rgb]{0,.555,.976}{\textbf{0.895}} & \textcolor{red}{\textbf{0.944}} & \textcolor{red}{\textbf{0.933}} & \textcolor[rgb]{0,.555,.976}{\textbf{0.905}} & \textcolor{red}{\textbf{0.926}} & \textcolor{red}{\textbf{1.000}} & \textcolor{red}{\textbf{0.957}} & \textcolor{red}{\textbf{1.000}} & \textcolor{red}{\textbf{1.000}} & \textcolor{red}{\textbf{0.981}} \\
			\midrule
			
			\midrule
			\textbf{Method} & \textbf{helmet0} & \textbf{helmet1} & \textbf{helmet2} & \textbf{helmet3} & \textbf{jar0} & \textbf{micro.0} & \textbf{shelf0} & \textbf{tap0} & \textbf{tap1} & \textbf{vase0} & \textbf{vase1} \\
			\midrule
			\textbf{BTF(Raw) (CVPR23’)} & 0.553 & 0.349 & 0.602 & 0.526 & 0.420 & 0.563 & 0.164 & 0.525 & 0.573 & 0.531 & 0.549 \\
			\textbf{BTF(FPFH) (CVPR23’)} & 0.571 & 0.719 & 0.542 & 0.444 & 0.424 & 0.671 & 0.609 & 0.560 & 0.546 & 0.342 & 0.219 \\
			\textbf{M3DM (CVPR23’)} & 0.526 & 0.427 & 0.623 & 0.374 & 0.441 & 0.357 & 0.564 & 0.754 & 0.739 & 0.423 & 0.427 \\
			\textbf{PatchCore(FPFH) (CVPR22’)} & 0.546 & 0.484 & 0.425 & 0.404 & 0.472 & 0.388 & 0.494 & 0.753 & 0.766 & 0.455 & 0.423 \\
			\textbf{PatchCore(PMAE) (CVPR22')} & 0.556 & 0.552 & 0.447 & 0.424 & 0.483 & 0.488 & 0.523 & 0.458 & 0.538 & 0.447 & 0.552 \\
			\textbf{CPMF (PR24’)} & 0.555 & 0.589 & 0.462 & 0.520 & 0.610 & 0.509 & 0.685 & 0.359 & 0.697 & 0.451 & 0.345 \\
			\textbf{Reg3D-AD (NeurIPS23’)} & 0.600 & 0.381 & 0.614 & 0.367 & 0.592 & 0.414 & 0.688 & 0.676 & 0.641 & 0.533 & 0.702 \\
			\textbf{IMRNet (CVPR24’)} & 0.597 & 0.600 & 0.641 & 0.573 & 0.780 & 0.755 & 0.603 & 0.676 & 0.696 & 0.533 & 0.757 \\
			\textbf{R3D-AD (ECCV24’)} & 0.757 & 0.720 & 0.633 & 0.707 & 0.838 & 0.762 & 0.696 & 0.736 & \textcolor[rgb]{0,.555,.976}{\textbf{0.900}} & 0.788 & 0.729 \\
			\textbf{PO3AD(CVPR25')} & 0.762 & 0.961 & 0.869 & 0.754 & 0.866 & 0.776 & 0.573 & 0.745 & 0.681 & 0.858 & 0.742 \\
			\textbf{MC3D-AD (IJCAI25’)} & 0.672 & \textcolor{red}{\textbf{1.000}} & 0.609 & \textcolor{red}{\textbf{0.979}} & \textcolor[rgb]{0,.555,.976}{\textbf{0.971}} & \textcolor[rgb]{0,.555,.976}{\textbf{0.919}} & \textcolor{red}{\textbf{0.841}} & \textcolor[rgb]{0,.555,.976}{\textbf{0.945}} & \textcolor{red}{\textbf{0.970}} & 0.821 & \textcolor[rgb]{0,.555,.976}{\textbf{0.857}} \\
			\textbf{Reg2Inv(NeurIPS25)} & \textcolor{red}{\textbf{0.817}} & \textcolor[rgb]{0,.555,.976}{\textbf{0.986}} & \textcolor[rgb]{0,.555,.976}{\textbf{0.875}} & \textcolor[rgb]{0,.555,.976}{\textbf{0.876}} & \textcolor{red}{\textbf{1.000}} & \textcolor{red}{\textbf{1.000}} & 0.577 & \textcolor{red}{\textbf{0.948}} & 0.804 & \textcolor{red}{\textbf{0.996}} & 0.605 \\
			\textbf{Simple3D (AAAI26’)} & 0.690 & 0.719 & 0.754 & 0.655 & 0.905 & \textcolor{red}{\textbf{1.000}} & \textcolor[rgb]{0,.555,.976}{\textbf{0.748}} & 0.703 & 0.596 & 0.954 & 0.824 \\
			\rowcolor{gray!10}\textbf{PA3AD(Ours)} & \textcolor[rgb]{0,.555,.976}{\textbf{0.803}} & \textcolor{red}{\textbf{1.000}} & \textcolor{red}{\textbf{0.907}} & \textcolor[rgb]{0,.555,.976}{\textbf{0.876}} & 0.952 & 0.862 & 0.704 & 0.924 & 0.782 & \textcolor[rgb]{0,.555,.976}{\textbf{0.975}} & \textcolor{red}{\textbf{0.905}} \\
			\midrule
			
			\midrule
			\textbf{Method} & \textbf{vase2} & \textbf{vase3} & \textbf{vase4} & \textbf{vase5} & \textbf{vase7} & \textbf{vase8} & \textbf{vase9} & \textbf{Average} & \textbf{Mean Rank} & \textbf{} & \textbf{} \\
			\midrule
			\textbf{BTF(Raw) (CVPR23’)} & 0.410 & 0.717 & 0.425 & 0.585 & 0.448 & 0.424 & 0.564 & 0.493 & 11.700 & & \\
			\textbf{BTF(FPFH) (CVPR23’)} & 0.546 & 0.699 & 0.510 & 0.409 & 0.518 & 0.668 & 0.268 & 0.528 & 10.975 & & \\
			\textbf{M3DM (CVPR23’)} & 0.737 & 0.439 & 0.476 & 0.317 & 0.657 & 0.663 & 0.663 & 0.552 & 10.675 & & \\
			\textbf{PatchCore(FPFH) (CVPR22’)} & 0.721 & 0.449 & 0.506 & 0.417 & 0.693 & 0.662 & 0.660 & 0.568 & 10.250 & & \\
			\textbf{PatchCore(PMAE) (CVPR22')} & 0.741 & 0.460 & 0.516 & 0.579 & 0.650 & 0.663 & 0.629 & 0.562 & 10.350 & & \\
			\textbf{CPMF (PR24’)} & 0.582 & 0.582 & 0.514 & 0.618 & 0.397 & 0.529 & 0.609 & 0.559 & 10.275 & & \\
			\textbf{Reg3D-AD (NeurIPS23’)} & 0.605 & 0.650 & 0.500 & 0.520 & 0.462 & 0.620 & 0.594 & 0.572 & 10.300 & & \\
			\textbf{IMRNet (CVPR24’)} & 0.614 & 0.700 & 0.524 & 0.676 & 0.635 & 0.630 & 0.594 & 0.661 & 7.750 & & \\
			\textbf{R3D-AD (ECCV24’)} & 0.752 & 0.742 & 0.630 & 0.757 & 0.771 & 0.721 & 0.718 & 0.749 & 5.600 & & \\
			\textbf{PO3AD(CVPR25')} & 0.952 & 0.821 & 0.675 & 0.852 & \textcolor[rgb]{0,.555,.976}{\textbf{0.966}} & 0.739 & 0.830 & 0.839 & 4.100 & & \\
			\textbf{MC3D-AD (IJCAI25’)} & 0.929 & 0.761 & \textcolor{red}{\textbf{0.876}} & 0.976 & 0.938 & 0.670 & 0.736 & 0.842 & 3.900 & & \\
			\textbf{Reg2Inv(NeurIPS25)} & \textcolor{red}{\textbf{1.000}} & \textcolor[rgb]{0,.555,.976}{\textbf{0.845}} & 0.818 & \textcolor{red}{\textbf{1.000}} & 0.643 & 0.818 & \textcolor[rgb]{0,.555,.976}{\textbf{0.873}} & \textcolor[rgb]{0,.555,.976}{\textbf{0.861}} & 3.425 &  &  \\
			\textbf{Simple3D (AAAI26’)} & 0.871 & 0.812 & \textcolor[rgb]{0,.555,.976}{\textbf{0.864}} & 0.962 & 0.895 & \textcolor[rgb]{0,.555,.976}{\textbf{0.855}} & 0.815 & 0.860 &\textcolor[rgb]{0,.555,.976}{\textbf{3.375}} &  &  \\
			\rowcolor{gray!10}\textbf{PA3AD(Ours)} & \textcolor[rgb]{0,.555,.976}{\textbf{0.981}} & \textcolor{red}{\textbf{0.918}} & 0.782 & \textcolor[rgb]{0,.555,.976}{\textbf{0.981}} & \textcolor{red}{\textbf{1.000}} & \textcolor{red}{\textbf{0.933}} & \textcolor{red}{\textbf{0.984}} & \textcolor{red}{\textbf{0.933}} & \textcolor{red}{\textbf{1.775}} & & \\
			\bottomrule
		\end{tabular}%
	}
\end{table*}

\begin{table*}[!ht]
	\centering
	\normalsize
	\setlength{\tabcolsep}{2.0pt}
	\renewcommand{\arraystretch}{0.70}
	\caption{P-AUROC performance of different methods on Anomaly-ShapeNet across 40 categories, where best and second-place results are highlighted in \textcolor[rgb]{1, 0, 0}{\textbf{red}} and \textcolor[rgb]{ 0,  .555,  .976}{\textbf{blue}}, respectively.}
	\label{p-auroc-anomalyshapenet}
	\resizebox{1\textwidth}{!}{
		\begin{tabular}{l|ccccccccccc}
			\toprule
			\multicolumn{12}{c}{\textbf{P-AUROC}} \\
			\midrule
			\textbf{Method} & \textbf{ashtray0} & \textbf{bag0} & \textbf{bottle0} & \textbf{bottle1} & \textbf{bottle3} & \textbf{bowl0} & \textbf{bowl1} & \textbf{bowl2} & \textbf{bowl3} & \textbf{bowl4} & \textbf{bowl5} \\
			\midrule
			\textbf{BTF(Raw)(CVPR23')} & 0.512 & 0.430 & 0.551 & 0.491 & 0.720 & 0.524 & 0.464 & 0.426 & 0.685 & 0.563 & 0.517 \\
			\textbf{BTF(FPFH)(CVPR23')} & 0.624 & 0.746 & 0.641 & 0.549 & 0.622 & 0.710 & 0.768 & 0.518 & 0.590 & 0.679 & 0.699 \\
			\textbf{M3DM(CVPR23')} & 0.577 & 0.637 & 0.663 & 0.637 & 0.532 & 0.658 & 0.663 & 0.694 & 0.657 & 0.624 & 0.489 \\
			\textbf{PatchCore(FPFH)(CVPR22')} & 0.597 & 0.574 & 0.654 & 0.687 & 0.512 & 0.524 & 0.531 & 0.625 & 0.327 & 0.720 & 0.358 \\
			\textbf{PatchCore(PointMAE)(CVPR22')} & 0.495 & 0.674 & 0.553 & 0.606 & 0.653 & 0.527 & 0.524 & 0.515 & 0.581 & 0.501 & 0.562 \\
			\textbf{CPMF(PR24')} & 0.615 & 0.655 & 0.521 & 0.571 & 0.435 & 0.745 & 0.488 & 0.635 & 0.641 & 0.683 & 0.684 \\
			\textbf{Reg3D-AD(NeurIPS23')} & 0.698 & 0.715 & 0.886 & 0.696 & 0.525 & 0.775 & 0.615 & 0.593 & 0.654 & 0.800 & 0.691 \\
			\textbf{IMRNet(CVPR24')} & 0.671 & 0.668 & 0.556 & 0.702 & 0.641 & 0.781 & 0.705 & 0.684 & 0.599 & 0.576 & 0.715 \\
			\textbf{ISMP(AAAI25')} & 0.865 & 0.734 & 0.722 & \textcolor[rgb]{ 0,  .555,  .976}{\textbf{0.869}} & 0.740 & 0.762 & 0.702 & 0.706 & 0.851 & 0.753 & 0.733 \\
			\textbf{RIF(PR25')} & 0.865 & 0.957 & 0.952 & 0.710 & 0.813 & 0.979 & 0.613 & 0.859 & 0.872 & 0.802 & 0.927 \\
			\textbf{PO3AD(CVPR25')} & \textcolor[rgb]{ 0,  .555,  .976}{\textbf{0.962}} & 0.949 & 0.912 & 0.844 & \textcolor[rgb]{ 0,  .555,  .976}{\textbf{0.880}} & 0.978 & \textcolor[rgb]{ 0,  .555,  .976}{\textbf{0.914}} & 0.918 & 0.935 & \textcolor[rgb]{ 0,  .555,  .976}{\textbf{0.967}} & 0.941 \\
			\textbf{Reg2Inv(NeurIPS25)} & 0.785 & \textcolor{red}{\textbf{0.991}} & \textcolor{red}{\textbf{0.995}} & 0.849 & 0.817 & \textcolor[rgb]{ 0,  .555,  .976}{\textbf{0.983}} & 0.828 & 0.822 & 0.761 & 0.788 & 0.824 \\
			\textbf{Simple3D (AAAI26’)} & 0.920 & 0.954 & \textcolor[rgb]{ 0,  .555,  .976}{\textbf{0.974}} & 0.728 & 0.838 & \textcolor{red}{\textbf{0.988}} & \textcolor{red}{\textbf{0.951}} & \textcolor[rgb]{ 0,  .555,  .976}{\textbf{0.933}} &  \textcolor{red}{\textbf{0.993}} & 0.929 & \textcolor{red}{\textbf{0.979}} \\
			\rowcolor{gray!10}\textbf{PA3AD(Ours)} & \textcolor{red}{\textbf{0.963}} & \textcolor[rgb]{ 0,  .555,  .976}{\textbf{0.965}} & 0.958 & \textcolor{red}{\textbf{0.915}} & \textcolor{red}{\textbf{0.940}} & 0.971 & 0.883 & \textcolor{red}{\textbf{0.934}} & \textcolor[rgb]{ 0,  .555,  .976}{\textbf{0.964}} & \textcolor{red}{\textbf{0.984}} & \textcolor[rgb]{ 0,  .555,  .976}{\textbf{0.948}} \\
			\midrule
			
			\midrule
			\textbf{Method} & \textbf{bucket0} & \textbf{bucket1} & \textbf{cap0} & \textbf{cap3} & \textbf{cap4} & \textbf{cap5} & \textbf{cup0} & \textbf{cup1} & \textbf{eraser0} & \textbf{headset0} & \textbf{headset1} \\
			\midrule
			\textbf{BTF(Raw)(CVPR23’)} & 0.617 & 0.686 & 0.524 & 0.687 & 0.469 & 0.373 & 0.632 & 0.561 & 0.637 & 0.578 & 0.475 \\
			\textbf{BTF(FPFH)(CVPR23’)} & 0.401 & 0.633 & 0.730 & 0.658 & 0.524 & 0.586 & 0.790 & 0.619 & 0.719 & 0.620 & 0.591 \\
			\textbf{M3DM(CVPR23’)} & 0.698 & 0.699 & 0.531 & 0.605 & 0.718 & 0.655 & 0.715 & 0.556 & 0.710 & 0.581 & 0.585 \\
			\textbf{PatchCore(FPFH)(CVPR22’)} & 0.459 & 0.571 & 0.472 & 0.653 & 0.595 & 0.795 & 0.655 & 0.596 & 0.810 & 0.583 & 0.464 \\
			\textbf{PatchCore(PointMAE)(CVPR22’)} & 0.586 & 0.574 & 0.544 & 0.488 & 0.725 & 0.545 & 0.510 & 0.856 & 0.378 & 0.575 & 0.423 \\
			\textbf{CPMF(PR24’)} & 0.486 & 0.601 & 0.601 & 0.551 & 0.553 & 0.551 & 0.497 & 0.509 & 0.689 & 0.699 & 0.458 \\
			\textbf{Reg3D-AD(NeurIPS23’)} & 0.619 & 0.752 & 0.632 & 0.718 & 0.815 & 0.467 & 0.685 & 0.698 & 0.755 & 0.580 & 0.626 \\
			\textbf{IMRNet(CVPR24’)} & 0.585 & 0.774 & 0.715 & 0.706 & 0.753 & 0.742 & 0.643 & 0.688 & 0.548 & 0.705 & 0.476 \\
			\textbf{ISMP(AAAI25')} & 0.545 & 0.683 & 0.672 & 0.775 & 0.661 & 0.770 & 0.552 & 0.851 & 0.524 & 0.472 & 0.843 \\
			\textbf{RIF(PR25')} & 0.553 & 0.812 & \textcolor[rgb]{ 0,  .555,  .976}{\textbf{0.968}} & \textcolor[rgb]{ 0,  .555,  .976}{\textbf{0.955}} & \textcolor[rgb]{ 0,  .555,  .976}{\textbf{0.956}} & 0.943 & 0.912 & 0.831 & \textcolor[rgb]{ 0,  .555,  .976}{\textbf{0.976}} & 0.895 & 0.840 \\
			\textbf{PO3AD(CVPR25')} & \textcolor[rgb]{ 0,  .555,  .976}{\textbf{0.755}} & 0.899 & 0.957 & 0.948 & 0.940 & 0.864 & 0.909 & 0.932 & 0.974 & 0.823 & 0.907 \\
			\textbf{Reg2Inv(NeurIPS25)} & 0.610 & 0.855 & 0.861 & 0.945 & 0.864 & \textcolor{red}{\textbf{0.970}} & 0.798 & 0.881 & \textcolor{red}{\textbf{0.980}} & \textcolor{red}{\textbf{0.946}} & \textcolor{red}{\textbf{0.970}} \\
			\textbf{Simple3D (AAAI26’)} & 0.725 & \textcolor{red}{\textbf{0.921}} & \textcolor{red}{\textbf{0.988}} & \textcolor{red}{\textbf{0.964}} & \textcolor{red}{\textbf{0.979}} & \textcolor[rgb]{ 0,  .555,  .976}{\textbf{0.964}} & \textcolor{red}{\textbf{0.979}} & \textcolor[rgb]{ 0,  .555,  .976}{\textbf{0.937}} & 0.970 & \textcolor[rgb]{ 0,  .555,  .976}{\textbf{0.936}} & \textcolor[rgb]{ 0,  .555,  .976}{\textbf{0.958}} \\
			\rowcolor{gray!10}\textbf{PA3AD(Ours)} & \textcolor{red}{\textbf{0.900}} & \textcolor[rgb]{ 0,  .555,  .976}{\textbf{0.920}} & 0.959 & \textcolor[rgb]{ 0,  .555,  .976}{\textbf{0.955}} & 0.945 & 0.963 & \textcolor[rgb]{ 0,  .555,  .976}{\textbf{0.975}} & \textcolor{red}{\textbf{0.958}} & \textcolor{red}{\textbf{0.980}} & 0.861 & 0.922 \\
			\midrule
			
			\midrule
			\textbf{Method} & \textbf{helmet0} & \textbf{helmet1} & \textbf{helmet2} & \textbf{helmet3} & \textbf{jar0} & \textbf{micro.0} & \textbf{shelf0} & \textbf{tap0} & \textbf{tap1} & \textbf{vase0} & \textbf{vase1} \\
			\midrule
			\textbf{BTF(Raw)(CVPR23’)} & 0.504 & 0.449 & 0.605 & 0.700 & 0.423 & 0.583 & 0.464 & 0.527 & 0.564 & 0.618 & 0.549 \\
			\textbf{BTF(FPFH)(CVPR23’)} & 0.575 & 0.749 & 0.643 & 0.724 & 0.427 & 0.675 & 0.619 & 0.568 & 0.596 & 0.642 & 0.619 \\
			\textbf{M3DM(CVPR23’)} & 0.599 & 0.427 & 0.623 & 0.655 & 0.541 & 0.358 & 0.554 & 0.654 & 0.712 & 0.608 & 0.602 \\
			\textbf{PatchCore(FPFH)(CVPR22’)} & 0.548 & 0.489 & 0.455 & 0.737 & 0.478 & 0.488 & 0.613 & 0.733 & 0.768 & 0.655 & 0.453 \\
			\textbf{PatchCore(PointMAE)(CVPR22’)} & 0.580 & 0.562 & 0.651 & 0.615 & 0.487 & 0.886 & 0.543 & 0.858 & 0.541 & 0.677 & 0.551 \\
			\textbf{CPMF(PR24’)} & 0.555 & 0.542 & 0.515 & 0.520 & 0.611 & 0.545 & \textcolor[rgb]{ 0,  .555,  .976}{\textbf{0.783}} & 0.458 & 0.657 & 0.458 & 0.486 \\
			\textbf{Reg3D-AD(NeurIPS23’)} & 0.600 & 0.624 & 0.825 & 0.620 & 0.599 & 0.599 & 0.688 & 0.589 & 0.741 & 0.548 & 0.602 \\
			\textbf{IMRNet(CVPR24’)} & 0.598 & 0.604 & 0.644 & 0.663 & 0.765 & 0.742 & 0.605 & 0.681 & 0.699 & 0.535 & 0.685 \\
			\textbf{ISMP(AAAI25')} & 0.615 & 0.603 & 0.568 & 0.522 & 0.661 & 0.600 & 0.701 & 0.844 & 0.678 & 0.687 & 0.534 \\
			\textbf{RIF(PR25')} & 0.838 & 0.639 & 0.905 & 0.719 & 0.967 & 0.955 & 0.765 & 0.670 & 0.667 & 0.905 & 0.807 \\
			\textbf{PO3AD(CVPR25')} & 0.878 & \textcolor[rgb]{ 0,  .555,  .976}{\textbf{0.948}} & 0.932 & 0.846 & 0.871 & 0.810 & 0.663 & 0.783 & 0.692 & 0.955 & \textcolor[rgb]{ 0,  .555,  .976}{\textbf{0.882}} \\
			\textbf{Reg2Inv(NeurIPS25)} & \textcolor{red}{\textbf{0.925}} & 0.906 & 0.891 & \textcolor{red}{\textbf{0.956}} & \textcolor{red}{\textbf{0.982}} & \textcolor{red}{\textbf{0.992}} & 0.632 & \textcolor{red}{\textbf{0.918}} & \textcolor{red}{\textbf{0.869}} & \textcolor{red}{\textbf{0.980}} & 0.705 \\
			\textbf{Simple3D (AAAI26’)} & \textcolor[rgb]{ 0,  .555,  .976}{\textbf{0.902}} & 0.902 & \textcolor[rgb]{ 0,  .555,  .976}{\textbf{0.947}} & \textcolor[rgb]{ 0,  .555,  .976}{\textbf{0.928}} & \textcolor[rgb]{ 0,  .555,  .976}{\textbf{0.978}} & \textcolor[rgb]{ 0,  .555,  .976}{\textbf{0.964}} & \textcolor{red}{\textbf{0.901}} & 0.857 & 0.811 & 0.934 & 0.807 \\
			\rowcolor{gray!10}\textbf{PA3AD(Ours)} & 0.846 & \textcolor{red}{\textbf{0.961}} & \textcolor{red}{\textbf{0.952}} & 0.899 & 0.953 & 0.879 & 0.780 & \textcolor[rgb]{ 0,  .555,  .976}{\textbf{0.894}} & \textcolor[rgb]{ 0,  .555,  .976}{\textbf{0.817}} & \textcolor[rgb]{ 0,  .555,  .976}{\textbf{0.974}} & \textcolor{red}{\textbf{0.939}} \\
			\midrule
			
			\midrule
			\textbf{Method} & \textbf{vase2} & \textbf{vase3} & \textbf{vase4} & \textbf{vase5} & \textbf{vase7} & \textbf{vase8} & \textbf{vase9} & \textbf{Average} & \textbf{Mean Rank} & \textbf{} & \textbf{} \\
			\midrule
			\textbf{BTF(Raw)(CVPR23’)} & 0.403 & 0.602 & 0.613 & 0.585 & 0.578 & 0.550 & 0.564 & 0.550 & 11.625 & & \\
			\textbf{BTF(FPFH)(CVPR23’)} & 0.646 & 0.699 & 0.710 & 0.429 & 0.540 & 0.662 & 0.568 & 0.628 & 9.450 & & \\
			\textbf{M3DM(CVPR23’)} & 0.737 & 0.658 & 0.655 & 0.642 & 0.517 & 0.551 & 0.663 & 0.616 & 9.750 & & \\
			\textbf{PatchCore(FPFH)(CVPR22’)} & 0.721 & 0.430 & 0.505 & 0.447 & 0.693 & 0.575 & 0.663 & 0.580 & 10.725 & & \\
			\textbf{PatchCore(PointMAE)(CVPR22’)} & 0.742 & 0.465 & 0.523 & 0.572 & 0.651 & 0.364 & 0.423 & 0.577 & 10.875 & & \\
			\textbf{CPMF(PR24’)} & 0.582 & 0.582 & 0.514 & 0.651 & 0.504 & 0.529 & 0.545 & 0.573 & 11.375 & & \\
			\textbf{Reg3D-AD(NeurIPS23’)} & 0.405 & 0.511 & 0.755 & 0.624 & 0.881 & 0.811 & 0.694 & 0.668 & 8.200 & & \\
			\textbf{IMRNet(CVPR24’)} & 0.614 & 0.401 & 0.524 & 0.682 & 0.593 & 0.635 & 0.691 & 0.650 & 8.700 & & \\
			\textbf{ISMP(AAAI25')} & 0.773 & 0.622 & 0.546 & 0.580 & 0.747 & 0.736 & 0.823 & 0.691 & 7.850 & & \\
			\textbf{RIF(PR25')} & 0.930 & 0.863 & 0.842 & 0.696 & 0.913 & 0.938 & 0.775 & 0.845 & 4.750 & & \\
			\textbf{PO3AD(CVPR25')} & 0.978 & 0.884 & 0.902 & 0.937 & 0.982 & \textcolor[rgb]{ 0,  .555,  .976}{\textbf{0.950}} & \textcolor[rgb]{ 0,  .555,  .976}{\textbf{0.952}} & 0.898 & 3.650 & & \\
			\textbf{Reg2Inv(NeurIPS25)} & \textcolor{red}{\textbf{0.997}} & 0.844 & \textcolor[rgb]{ 0,  .555,  .976}{\textbf{0.927}} & 0.879 & 0.986 & 0.934 & \textcolor{red}{\textbf{0.971}} & 0.882 & 3.350 &  &  \\
			\textbf{Simple3D (AAAI26’)} & \textcolor[rgb]{ 0,  .555,  .976}{\textbf{0.986}} & \textcolor{red}{\textbf{0.910}} & \textcolor{red}{\textbf{0.985}} & \textcolor{red}{\textbf{0.966}} & \textcolor[rgb]{ 0,  .555,  .976}{\textbf{0.990}} & \textcolor{red}{\textbf{0.975}} & 0.913 & \textcolor[rgb]{ 0,  .555,  .976}{\textbf{0.929}} & \textcolor[rgb]{ 0,  .555,  .976}{\textbf{2.300}} &  &  \\
			\rowcolor{gray!10}\textbf{PA3AD(Ours)} & 0.983 & \textcolor[rgb]{ 0,  .555,  .976}{\textbf{0.889}} & 0.891 & \textcolor[rgb]{ 0,  .555,  .976}{\textbf{0.962}} & \textcolor{red}{\textbf{0.992}} & \textcolor[rgb]{ 0,  .555,  .976}{\textbf{0.950}} & 0.920 & \textcolor{red}{\textbf{0.931}} & \textcolor{red}{\textbf{2.200}} & & \\
			\bottomrule
		\end{tabular}
	}
\end{table*}

\begin{table*}[!ht]
	\centering
	\normalsize
	\setlength{\tabcolsep}{1.5pt}
	\renewcommand{\arraystretch}{0.70}
	\caption{O-AUPR performance of different methods on Anomaly-ShapeNet across 40 categories, where best and second-place results are highlighted in \textcolor[rgb]{1, 0, 0}{\textbf{red}} and \textcolor[rgb]{ 0,  .555,  .976}{\textbf{blue}}, respectively.}
	\label{o-aupr-anomalyshapenet}
	\resizebox{1\textwidth}{!}{
		\begin{tabular}{l|ccccccccccc}
			\toprule
			\multicolumn{12}{c}{\textbf{O-AUPR}} \\
			\midrule
			\textbf{Method} & \textbf{ashtray0} & \textbf{bag0} & \textbf{bottle0} & \textbf{bottle1} & \textbf{bottle3} & \textbf{bowl0} & \textbf{bowl1} & \textbf{bowl2} & \textbf{bowl3} & \textbf{bowl4} & \textbf{bowl5} \\
			\midrule
			\textbf{BTF(Raw)(CVPR23’)}      & 0.578 & 0.458 & 0.466 & 0.573 & 0.543 & 0.588 & 0.464 & 0.576 & 0.654 & 0.601 & 0.615 \\
			\textbf{BTF(FPFH)(CVPR23’)}     & 0.651 & 0.551 & 0.644 & 0.625 & 0.602 & 0.576 & 0.648 & 0.515 & 0.499 & 0.632 & 0.699 \\
			\textbf{M3DM(CVPR23’)}          & 0.632 & 0.642 & 0.763 & 0.674 & 0.451 & 0.525 & 0.515 & 0.630 & 0.635 & 0.571 & 0.601 \\
			\textbf{PatchCore(FPFH)(CVPR22’)} & 0.445 & 0.608 & 0.615 & 0.677 & 0.579 & 0.548 & 0.545 & 0.611 & 0.620 & 0.575 & 0.541 \\
			\textbf{PatchCore(PointMAE)(CVPR22’)} & 0.679 & 0.601 & 0.545 & 0.645 & 0.651 & 0.562 & 0.611 & 0.456 & 0.556 & 0.601 & 0.585 \\
			\textbf{CPMF(PR24’)}            & 0.453 & 0.655 & 0.588 & 0.592 & 0.505 & 0.775 & 0.621 & 0.601 & 0.418 & 0.683 & 0.685 \\
			\textbf{Reg3D-AD(NeurIPS23’)}   & 0.588 & 0.608 & 0.632 & 0.695 & 0.474 & 0.494 & 0.515 & 0.495 & 0.441 & 0.624 & 0.555 \\
			\textbf{IMRNet(CVPR24’)}        & 0.612 & 0.665 & 0.558 & 0.702 & 0.648 & 0.481 & 0.504 & 0.681 & 0.614 & 0.630 & 0.652 \\
			\textbf{PO3AD(CVPR25’)}         & \textcolor[rgb]{ 0,  .555,  .976}{\textbf{0.999}} & \textcolor[rgb]{ 0,  .555,  .976}{\textbf{0.809}} & \textcolor[rgb]{ 0,  .555,  .976}{\textbf{0.927}} & \textcolor[rgb]{ 0,  .555,  .976}{\textbf{0.959}} & \textcolor[rgb]{ 0,  .555,  .976}{\textbf{0.962}} & \textcolor[rgb]{ 0,  .555,  .976}{\textbf{0.946}} & \textcolor[rgb]{ 0,  .555,  .976}{\textbf{0.905}} & \textcolor[rgb]{ 0,  .555,  .976}{\textbf{0.888}} & \textcolor[rgb]{ 0,  .555,  .976}{\textbf{0.927}} & \textcolor[rgb]{ 0,  .555,  .976}{\textbf{0.985}} & \textcolor[rgb]{ 0,  .555,  .976}{\textbf{0.904}} \\
			\rowcolor{gray!10}\textbf{PA3AD(Ours)} & \textcolor{red}{\textbf{1.000}} & \textcolor{red}{\textbf{0.948}} & \textcolor{red}{\textbf{0.979}} & \textcolor{red}{\textbf{0.997}} & \textcolor{red}{\textbf{0.982}} & \textcolor{red}{\textbf{0.991}} & \textcolor{red}{\textbf{0.943}} & \textcolor{red}{\textbf{0.962}} & \textcolor{red}{\textbf{0.944}} & \textcolor{red}{\textbf{0.991}} & \textcolor{red}{\textbf{0.951}} \\
			\midrule
			
			\midrule
			\textbf{Method} & \textbf{bucket0} & \textbf{bucket1} & \textbf{cap0} & \textbf{cap3} & \textbf{cap4} & \textbf{cap5} & \textbf{cup0} & \textbf{cup1} & \textbf{eraser0} & \textbf{headset0} & \textbf{headset1} \\
			\midrule
			\textbf{BTF(Raw)(CVPR23’)}      & 0.652 & 0.620 & 0.659 & 0.612 & 0.515 & 0.653 & 0.601 & 0.701 & 0.425 & 0.379 & 0.515 \\
			\textbf{BTF(FPFH)(CVPR23’)}     & 0.483 & 0.648 & 0.618 & 0.579 & 0.545 & 0.593 & 0.585 & 0.651 & 0.719 & 0.531 & 0.523 \\
			\textbf{M3DM(CVPR23’)}          & 0.609 & 0.507 & 0.564 & 0.652 & 0.477 & 0.642 & 0.570 & 0.752 & 0.625 & 0.632 & 0.623 \\
			\textbf{PatchCore(FPFH)(CVPR22’)} & 0.604 & 0.565 & 0.585 & 0.457 & 0.655 & 0.725 & 0.604 & 0.586 & 0.584 & 0.701 & 0.601 \\
			\textbf{PatchCore(PointMAE)(CVPR22’)} & 0.541 & 0.642 & 0.561 & 0.583 & 0.721 & 0.542 & 0.642 & 0.710 & 0.801 & 0.515 & 0.423 \\
			\textbf{CPMF(PR24’)}            & 0.662 & 0.501 & 0.601 & 0.541 & 0.645 & 0.697 & 0.647 & 0.609 & 0.544 & 0.602 & 0.619 \\
			\textbf{Reg3D-AD(NeurIPS23’)}   & 0.632 & 0.714 & 0.693 & 0.711 & 0.623 & 0.770 & 0.531 & 0.638 & 0.424 & 0.538 & 0.617 \\
			\textbf{IMRNet(CVPR24’)}        & 0.578 & 0.732 & 0.711 & 0.702 & 0.658 & 0.502 & 0.455 & 0.627 & 0.599 & 0.701 & 0.656 \\
			\textbf{PO3AD(CVPR25’)}         & \textcolor[rgb]{ 0,  .555,  .976}{\textbf{0.923}} & \textcolor[rgb]{ 0,  .555,  .976}{\textbf{0.882}} & \textcolor[rgb]{ 0,  .555,  .976}{\textbf{0.841}} & \textcolor[rgb]{ 0,  .555,  .976}{\textbf{0.906}} & \textcolor[rgb]{ 0,  .555,  .976}{\textbf{0.876}} & \textcolor[rgb]{ 0,  .555,  .976}{\textbf{0.801}} & \textcolor[rgb]{ 0,  .555,  .976}{\textbf{0.879}} & \textcolor[rgb]{ 0,  .555,  .976}{\textbf{0.870}} & \textcolor[rgb]{ 0,  .555,  .976}{\textbf{0.995}} & \textcolor[rgb]{ 0,  .555,  .976}{\textbf{0.765}} & \textcolor[rgb]{ 0,  .555,  .976}{\textbf{0.914}} \\
			\rowcolor{gray!10}\textbf{PA3AD(Ours)} & \textcolor{red}{\textbf{0.982}} & \textcolor{red}{\textbf{0.937}} & \textcolor{red}{\textbf{0.967}} & \textcolor{red}{\textbf{0.952}} & \textcolor{red}{\textbf{0.935}} & \textcolor{red}{\textbf{0.946}} & \textcolor{red}{\textbf{1.000}} & \textcolor{red}{\textbf{0.964}} & \textcolor{red}{\textbf{1.000}} & \textcolor{red}{\textbf{1.000}} & \textcolor{red}{\textbf{0.982}} \\
			\midrule
			
			\midrule
			\textbf{Method} & \textbf{helmet0} & \textbf{helmet1} & \textbf{helmet2} & \textbf{helmet3} & \textbf{jar0} & \textbf{micro.0} & \textbf{shelf0} & \textbf{tap0} & \textbf{tap1} & \textbf{vase0} & \textbf{vase1} \\
			\midrule
			\textbf{BTF(Raw)(CVPR23’)}      & 0.559 & 0.388 & 0.615 & 0.526 & 0.428 & 0.613 & 0.624 & 0.535 & 0.594 & 0.562 & 0.441 \\
			\textbf{BTF(FPFH)(CVPR23’)}     & 0.568 & 0.721 & 0.588 & 0.564 & 0.479 & 0.662 & 0.611 & 0.610 & 0.575 & 0.641 & 0.655 \\
			\textbf{M3DM(CVPR23’)}          & 0.528 & 0.627 & 0.636 & 0.458 & 0.555 & 0.464 & 0.665 & 0.722 & 0.638 & \textcolor[rgb]{ 0,  .555,  .976}{\textbf{0.788}} & 0.652 \\
			\textbf{PatchCore(FPFH)(CVPR22’)} & 0.525 & 0.630 & 0.475 & 0.494 & 0.499 & 0.332 & 0.504 & 0.712 & 0.684 & 0.645 & 0.623 \\
			\textbf{PatchCore(PointMAE)(CVPR22’)} & 0.633 & 0.571 & 0.496 & 0.611 & 0.463 & 0.652 & 0.543 & 0.712 & 0.542 & 0.548 & 0.572 \\
			\textbf{CPMF(PR24’)}            & 0.333 & 0.501 & 0.477 & 0.645 & 0.618 & 0.655 & \textcolor[rgb]{ 0,  .555,  .976}{\textbf{0.681}} & 0.639 & 0.697 & 0.632 & 0.645 \\
			\textbf{Reg3D-AD(NeurIPS23’)}   & 0.600 & 0.381 & 0.618 & 0.468 & 0.601 & 0.614 & 0.675 & 0.676 & 0.599 & 0.615 & 0.468 \\
			\textbf{IMRNet(CVPR24’)}        & 0.697 & 0.615 & 0.602 & 0.575 & 0.760 & 0.552 & 0.625 & 0.401 & \textcolor[rgb]{ 0,  .555,  .976}{\textbf{0.796}} & 0.573 & 0.725 \\
			\textbf{PO3AD(CVPR25’)}         & \textcolor[rgb]{ 0,  .555,  .976}{\textbf{0.864}} & \textcolor[rgb]{ 0,  .555,  .976}{\textbf{0.961}} & \textcolor[rgb]{ 0,  .555,  .976}{\textbf{0.934}} & \textcolor[rgb]{ 0,  .555,  .976}{\textbf{0.849}} & \textcolor[rgb]{ 0,  .555,  .976}{\textbf{0.915}} & \textcolor[rgb]{ 0,  .555,  .976}{\textbf{0.803}} & 0.680 & \textcolor[rgb]{ 0,  .555,  .976}{\textbf{0.856}} & 0.709 & 0.753 & \textcolor[rgb]{ 0,  .555,  .976}{\textbf{0.789}} \\
			\rowcolor{gray!10}\textbf{PA3AD(Ours)} & \textcolor{red}{\textbf{0.885}} & \textcolor{red}{\textbf{1.000}} & \textcolor{red}{\textbf{0.957}} & \textcolor{red}{\textbf{0.935}} & \textcolor{red}{\textbf{0.954}} & \textcolor{red}{\textbf{0.813}} & \textcolor{red}{\textbf{0.824}} & \textcolor{red}{\textbf{0.958}} & \textcolor{red}{\textbf{0.820}} & \textcolor{red}{\textbf{0.996}} & \textcolor{red}{\textbf{0.921}} \\
			\midrule
			
			\midrule
			\textbf{Method} & \textbf{vase2} & \textbf{vase3} & \textbf{vase4} & \textbf{vase5} & \textbf{vase7} & \textbf{vase8} & \textbf{vase9} & \textbf{Average} & \textbf{Mean Rank} & \textbf{} & \textbf{} \\
			\midrule
			\textbf{BTF(Raw)(CVPR23’)}      & 0.413 & 0.717 & 0.428 & 0.615 & 0.547 & 0.416 & 0.482 & 0.549 & 7.525 & & \\
			\textbf{BTF(FPFH)(CVPR23’)}     & 0.569 & 0.652 & 0.587 & 0.472 & 0.592 & 0.624 & 0.638 & 0.598 & 6.350 & & \\
			\textbf{M3DM(CVPR23’)}          & 0.615 & 0.551 & 0.526 & 0.633 & 0.648 & 0.463 & 0.651 & 0.603 & 6.175 & & \\
			\textbf{PatchCore(FPFH)(CVPR22’)} & 0.801 & 0.481 & 0.777 & 0.515 & 0.621 & 0.515 & 0.660 & 0.588 & 6.675 & & \\
			\textbf{PatchCore(PointMAE)(CVPR22’)} & 0.711 & 0.455 & 0.586 & 0.585 & 0.652 & 0.655 & 0.634 & 0.595 & 6.625 & & \\
			\textbf{CPMF(PR24’)}            & 0.632 & 0.588 & 0.655 & 0.518 & 0.432 & 0.673 & 0.618 & 0.597 & 6.175 & & \\
			\textbf{Reg3D-AD(NeurIPS23’)}   & 0.641 & 0.651 & 0.505 & 0.588 & 0.455 & 0.629 & 0.574 & 0.584 & 6.675 & & \\
			\textbf{IMRNet(CVPR24’)}        & 0.655 & 0.708 & 0.528 & 0.654 & 0.601 & 0.639 & 0.462 & 0.621 & 5.600 & & \\
			\textbf{PO3AD(CVPR25’)}         & \textcolor[rgb]{ 0,  .555,  .976}{\textbf{0.963}} & \textcolor[rgb]{ 0,  .555,  .976}{\textbf{0.902}} & \textcolor[rgb]{ 0,  .555,  .976}{\textbf{0.824}} & \textcolor[rgb]{ 0,  .555,  .976}{\textbf{0.879}} & \textcolor[rgb]{ 0,  .555,  .976}{\textbf{0.971}} & \textcolor[rgb]{ 0,  .555,  .976}{\textbf{0.833}} & \textcolor[rgb]{ 0,  .555,  .976}{\textbf{0.904}} & \textcolor[rgb]{ 0,  .555,  .976}{\textbf{0.881}} & \textcolor[rgb]{ 0,  .555,  .976}{\textbf{2.075}} & & \\
			\rowcolor{gray!10}\textbf{PA3AD(Ours)} & \textcolor{red}{\textbf{0.984}} & \textcolor{red}{\textbf{0.952}} & \textcolor{red}{\textbf{0.861}} & \textcolor{red}{\textbf{0.982}} & \textcolor{red}{\textbf{1.000}} & \textcolor{red}{\textbf{0.964}} & \textcolor{red}{\textbf{0.931}} & \textcolor{red}{\textbf{0.952}} & \textcolor{red}{\textbf{1.000}} & & \\
			\bottomrule
		\end{tabular}
	}
\end{table*}

\subsection{Comparison methods}

We compare the proposed method with representative state-of-the-art approaches for 3D anomaly detection and localization on point clouds, including PatchCore\cite{roth2022towards}, BTF\cite{horwitz2023back}, M3DM\cite{wang2023multimodal}, CPMF\cite{cao2024complementary}, Reg3D-AD\cite{liu2023real3d}, IMR-Net\cite{li2024towards}, R3D-AD\cite{zhou2024r3d}, MC3D-AD\cite{cheng2025mc3d}, ISMP\cite{liang2025look}, PO3AD\cite{ye2025po3ad}, RIF\cite{LIANG2026112924}, Reg2Inv\cite{Reg2Inv}, and Simple3D\cite{MiniShift_Simple3D}. Overall, these baselines span a broad spectrum of design philosophies, including retrieval-based methods with classical 3D geometric cues, multimodal or hybrid-fusion schemes, memory bank based modeling for improved robustness, synthesis-based learning, reconstruction based distribution modeling, and regression-style formulations for fine-grained localization. PatchCore is originally proposed for 2D anomaly detection and is extended to 3D by replacing the feature extractor with a point-cloud encoder. For a fair comparison, we follow the standard evaluation protocol used in prior work and directly adopt the reported results of all competing methods from their original papers.

\subsection{Comparisons on Anomaly-ShapeNet dataset}
\label{section4.4}

%Table \ref{o-auroc-anomalyshapenet} and Table \ref{p-auroc-anomalyshapenet} report object-level and point-level results on Anomaly-ShapeNet across 40 classes and compare our method with recent state-of-the-art approaches. Experiments show that our PA3AD achieves the best performance on nearly all evaluation metrics, with substantial gains over prior methods. At the object level, our method attains 100\% detection across several classes. On average performance, it surpasses the current methods by 10.8\% at the object-level and by 3.3\% at the point-level, demonstrating the superiority of the proposed approach.

%Additionally, Table \ref{o-aupr-anomalyshapenet} presents object‑level AUPR results on Anomaly-ShapeNet compared with recent state‑of‑the‑art methods. Different from AUROC, AUPR is more sensitive to class imbalance and better reflects practical detection performance. Our method achieves the best results across all classes, attaining a mean AUPR of 95.2\%, which is 8\% higher than the current methods, further corroborating its effectiveness.

Table \ref{o-auroc-anomalyshapenet} and Table \ref{p-auroc-anomalyshapenet} report object-level and point-level results on Anomaly-ShapeNet across 40 classes. Our PA3AD achieves the best performance on nearly all metrics, attaining 100\% detection in several classes and surpassing prior methods by 8.4\% in object-level AUROC.

Additionally, Table \ref{o-aupr-anomalyshapenet} presents object-level AUPR results on Anomaly-ShapeNet. Unlike AUROC, AUPR is more sensitive to class imbalance and better reflects practical detection performance. Our method achieves a mean AUPR of 95.2\%, 8\% higher than existing methods, confirming that coupling physics-inspired pseudo-anomalies with the difference-aware offset head raises small-defect recall without sacrificing boundary precision.

\begin{table*}[!t]
	\centering
	\normalsize
	\setlength{\tabcolsep}{2.0pt}
	\renewcommand{\arraystretch}{0.65}
	\caption{The O-AUROC performance ($\uparrow$) of different methods on Real3D-AD across 12 categories, where best and second-place results are highlighted in \textcolor[rgb]{1, 0, 0}{\textbf{red}} and \textcolor[rgb]{ 0,  .555,  .976}{\textbf{blue}}, respectively.}
	\label{o-auroc-real3d}
	\resizebox{1\textwidth}{!}{
		\begin{tabular}{l|ccccccc}
			\toprule
			\multicolumn{8}{c}{\textbf{O-AUROC}} \\
			\midrule
			\textbf{Method} & \textbf{Airplane} & \textbf{Car} & \textbf{Candy} & \textbf{Chicken} & \textbf{Diamond} & \textbf{Duck} & \textbf{Fish} \\
			\midrule
			\textbf{BTF(Raw) (CVPR23')} & 0.730 & 0.647 & 0.539 & 0.789 & 0.707 & 0.691 & 0.602 \\
			\textbf{BTF(FPFH) (CVPR23')} & 0.520 & 0.560 & 0.630 & 0.432 & 0.545 & 0.784 & 0.549 \\
			\textbf{M3DM (CVPR23')} & 0.434 & 0.541 & 0.552 & 0.683 & 0.602 & 0.433 & 0.540 \\
			\textbf{PatchCore(FPFH) (CVPR22')} & \textcolor{red}{\textbf{0.882}} & 0.590 & 0.541 & \textcolor[rgb]{ 0,  .555,  .976}{\textbf{0.837}} & 0.574 & 0.546 & 0.675 \\
			\textbf{PatchCore(PointMAE) (CVPR22')} & 0.726 & 0.498 & 0.663 & 0.827 & 0.783 & 0.489 & 0.630 \\
			\textbf{CPMF (PR24')} & 0.701 & 0.551 & 0.552 & 0.504 & 0.523 & 0.582 & 0.558 \\
			\textbf{Reg3D-AD (NeurIPS23')} & 0.716 & 0.697 & 0.685 & \textcolor{red}{\textbf{0.852}} & 0.900 & 0.584 & 0.915 \\
			\textbf{IMRNet (CVPR24')} & 0.762 & 0.711 & 0.755 & 0.780 & 0.905 & 0.517 & 0.880 \\
			\textbf{R3D-AD (ECCV24')} & 0.772 & 0.696 & 0.713 & 0.714 & 0.685 & \textcolor{red}{\textbf{0.909}} & 0.692 \\
			\textbf{ISMP (AAAI25')} & \textcolor[rgb]{ 0,  .555,  .976}{\textbf{0.858}} & 0.731 & \textcolor{red}{\textbf{0.852}} & 0.714 & \textcolor[rgb]{ 0,  .555,  .976}{\textbf{0.948}} & 0.712 & \textcolor[rgb]{ 0,  .555,  .976}{\textbf{0.945}} \\
			\textbf{PO3AD (CVPR25')} & 0.804 & 0.654 & 0.785 & 0.686 & 0.801 & 0.820 & 0.859 \\
			\textbf{MC3D-AD (IJCAI25’)} & 0.850 & \textcolor[rgb]{ 0,  .555,  .976}{\textbf{0.749}} & \textcolor[rgb]{ 0,  .555,  .976}{\textbf{0.830}} & 0.715 & \textcolor{red}{\textbf{0.955}} & \textcolor[rgb]{ 0,  .555,  .976}{\textbf{0.831}} & 0.865 \\
			\textbf{PA3AD(Ours)} & 0.760 & \textcolor{red}{\textbf{0.769}} & 0.752 & 0.677 & 0.776 & 0.782 & \textcolor{red}{\textbf{0.954}} \\
			\midrule
			
			\midrule
			\textbf{Method} & \textbf{Gemstone} & \textbf{Seahorse} & \textbf{Shell} & \textbf{Starfish} & \textbf{Toffees} & \textbf{Average} & \textbf{Mean Rank} \\
			\midrule
			\textbf{BTF(Raw) (CVPR23')} & 0.686 & 0.596 & 0.396 & 0.530 & 0.703 & 0.635 & 8.417 \\
			\textbf{BTF(FPFH) (CVPR23')} & 0.648 & \textcolor[rgb]{ 0,  .555,  .976}{\textbf{0.779}} & 0.754 & 0.575 & 0.462 & 0.603 & 8.667 \\
			\textbf{M3DM (CVPR23')} & 0.644 & 0.495 & 0.694 & 0.551 & 0.450 & 0.552 & 10.667 \\
			\textbf{PatchCore(FPFH) (CVPR22')} & 0.370 & 0.505 & 0.589 & 0.441 & 0.565 & 0.593 & 9.250 \\
			\textbf{PatchCore(PointMAE) (CVPR22')} & 0.374 & 0.539 & 0.501 & 0.519 & 0.585 & 0.594 & 9.583 \\
			\textbf{CPMF (PR24')} & 0.589 & 0.729 & 0.653 & 0.700 & 0.390 & 0.586 & 9.667 \\
			\textbf{Reg3D-AD (NeurIPS23')} & 0.417 & 0.762 & 0.583 & 0.506 & \textcolor[rgb]{ 0,  .555,  .976}{\textbf{0.827}} & 0.704 & 6.417 \\
			\textbf{IMRNet (CVPR24')} & 0.674 & 0.604 & 0.665 & 0.674 & 0.774 & 0.725 & 5.583 \\
			\textbf{R3D-AD (ECCV24')} & 0.665 & 0.720 & \textcolor{red}{\textbf{0.840}} & 0.701 & 0.703 & 0.734 & 5.417 \\
			\textbf{ISMP (AAAI25')} & 0.468 & 0.729 & 0.623 & 0.660 & \textcolor{red}{\textbf{0.842}} & 0.757 & 4.583 \\
			\textbf{PO3AD (CVPR25')} & \textcolor[rgb]{ 0,  .555,  .976}{\textbf{0.693}} & 0.756 & 0.800 & 0.758 & 0.771 & 0.765 & 4.583 \\
			\textbf{MC3D-AD (IJCAI25’)} & 0.560 & 0.716 & 0.803 & \textcolor[rgb]{ 0,  .555,  .976}{\textbf{0.766}} & 0.738 & \textcolor[rgb]{ 0,  .555,  .976}{\textbf{0.782}} & \textcolor[rgb]{ 0,  .555,  .976}{\textbf{4.083}} \\
			\textbf{PA3AD(Ours)} & \textcolor{red}{\textbf{0.728}} & \textcolor{red}{\textbf{0.875}} & \textcolor[rgb]{ 0,  .555,  .976}{\textbf{0.808}} & \textcolor{red}{\textbf{0.796}} & 0.789 & \textcolor{red}{\textbf{0.789}} & \textcolor{red}{\textbf{3.750}} \\
			\bottomrule
		\end{tabular}%
	}
\end{table*}

\begin{table*}[t]
	\centering
	\normalsize
	\setlength{\tabcolsep}{2pt}
	\renewcommand{\arraystretch}{0.65}
	\caption{P-AUROC performance ($\uparrow$) of different methods on Real3D-AD across 12 categories, where best and second-place results are highlighted in \textcolor[rgb]{1, 0, 0}{\textbf{red}} and \textcolor[rgb]{ 0,  .555,  .976}{\textbf{blue}}, respectively.}
	\label{p-auroc-real3d}
	\resizebox{1\textwidth}{!}{
		\begin{tabular}{l|ccccccc}
			\toprule
			\textbf{Method} & \textbf{Airplane} & \textbf{Car} & \textbf{Candy} & \textbf{Chicken} & \textbf{Diamond} & \textbf{Duck} & \textbf{Fish} \\
			\midrule
			\textbf{BTF(Raw) (CVPR23')} & 0.564 & 0.647 & 0.735 & 0.609 & 0.563 & 0.601 & 0.514 \\
			\textbf{BTF(FPFH) (CVPR23')} & 0.738 & 0.708 & 0.864 & 0.735 & 0.882 & 0.875 & 0.709 \\
			\textbf{M3DM (CVPR23')} & 0.547 & 0.602 & 0.679 & 0.678 & 0.608 & 0.667 & 0.606 \\
			\textbf{PatchCore(FPFH) (CVPR22')} & 0.562 & 0.754 & 0.780 & 0.429 & 0.828 & 0.264 & 0.829 \\
			\textbf{PatchCore(PointMAE) (CVPR22')} & 0.569 & 0.609 & 0.627 & 0.729 & 0.718 & 0.528 & 0.717 \\
			\textbf{Reg3D-AD (NeurIPS23')} & 0.631 & 0.718 & 0.724 & 0.676 & 0.835 & 0.503 & 0.826 \\
			\textbf{R3D-AD (ECCV24')} & 0.594 & 0.557 & 0.593 & 0.620 & 0.555 & 0.635 & 0.573 \\
			\textbf{ISMP (AAAI25')} & \textcolor{red}{\textbf{0.753}} & \textcolor[rgb]{ 0,  .555,  .976}{\textbf{0.836}} & \textcolor[rgb]{ 0,  .555,  .976}{\textbf{0.907}} & \textcolor[rgb]{ 0,  .555,  .976}{\textbf{0.798}} & \textcolor[rgb]{ 0,  .555,  .976}{\textbf{0.926}} & \textcolor[rgb]{ 0,  .555,  .976}{\textbf{0.876}} & 0.886 \\
			\textbf{MC3D-AD (IJCAI25')} & 0.628 & 0.819 & \textcolor{red}{\textbf{0.910}} & 0.640 & \textcolor{red}{\textbf{0.942}} & 0.822 & \textcolor{red}{\textbf{0.932}} \\
			\textbf{RIF(PR25')} & 0.733 & 0.777 & 0.913 & 0.728 & 0.975 & 0.869 & 0.710 \\
			\textbf{PA3AD (Ours)} & \textcolor[rgb]{ 0,  .555,  .976}{\textbf{0.742}} & \textcolor{red}{\textbf{0.880}} & 0.749 & \textcolor{red}{\textbf{0.890}} & 0.861 & \textcolor{red}{\textbf{0.878}} & \textcolor[rgb]{ 0,  .555,  .976}{\textbf{0.910}} \\
			\midrule
			\textbf{Method} & \textbf{Gemstone} & \textbf{Seahorse} & \textbf{Shell} & \textbf{Starfish} & \textbf{Toffees} & \textbf{Average} & \textbf{Mean Rank} \\
			\midrule
			\textbf{BTF(Raw) (CVPR23')} & 0.597 & 0.520 & 0.489 & 0.392 & 0.623 & 0.571 & 9.333 \\
			\textbf{BTF(FPFH) (CVPR23')} & \textcolor[rgb]{ 0,  .555,  .976}{\textbf{0.891}} & 0.512 & 0.571 & 0.501 & 0.815 & 0.733 & 5.917 \\
			\textbf{M3DM (CVPR23')} & 0.674 & 0.560 & 0.738 & 0.532 & 0.682 & 0.631 & 8.167 \\
			\textbf{PatchCore(FPFH) (CVPR22')} & \textcolor{red}{\textbf{0.910}} & 0.739 & 0.739 & 0.606 & 0.747 & 0.682 & 6.417 \\
			\textbf{PatchCore(PointMAE) (CVPR22')} & 0.444 & 0.633 & 0.709 & 0.580 & 0.580 & 0.620 & 8.083 \\
			\textbf{Reg3D-AD (NeurIPS23')} & 0.545 & \textcolor[rgb]{ 0,  .555,  .976}{\textbf{0.817}} & 0.811 & 0.617 & 0.759 & 0.705 & 5.917 \\
			\textbf{R3D-AD (ECCV24')} & 0.668 & 0.562 & 0.578 & 0.608 & 0.568 & 0.592 & 8.833 \\
			\textbf{ISMP (AAAI25')} & 0.857 & 0.813 & \textcolor[rgb]{ 0,  .555,  .976}{\textbf{0.839}} & 0.641 & \textcolor[rgb]{ 0,  .555,  .976}{\textbf{0.895}} & \textcolor[rgb]{ 0,  .555,  .976}{\textbf{0.836}} & \textcolor{red}{\textbf{2.500}} \\
			\textbf{MC3D-AD (IJCAI25')} & 0.458 & 0.659 & 0.778 & \textcolor[rgb]{ 0,  .555,  .976}{\textbf{0.690}} & \textcolor{red}{\textbf{0.934}} & 0.768 & \textcolor[rgb]{ 0,  .555,  .976}{\textbf{4.083}} \\
			\textbf{RIF(PR25')} & 0.935 & 0.573 & 0.714 & 0.588 & 0.888 & 0.784 & 4.250 \\
			\textbf{PA3AD (Ours)} & 0.787 & \textcolor{red}{\textbf{0.885}} & \textcolor{red}{\textbf{0.880}} & \textcolor{red}{\textbf{0.768}} & 0.832 & \textcolor{red}{\textbf{0.839}} & \textcolor{red}{\textbf{2.500}} \\
			\bottomrule
		\end{tabular}%
	}
\end{table*}

\subsection{Comparisons on Real3D-AD dataset}
\label{section4.5}

Table \ref{o-auroc-real3d} and Table \ref{p-auroc-real3d} report object-level and point-level results on the Real3D-AD dataset, showing that our method attains strong performance on the majority of categories. Strong gains are observed on classes with long-range structure or curved geometry, such as the object-level and point-level AUROC for fish and seahorse, whereas categories like airplane still leave room for improvement. These improvements are largely attributable to the local-global attention, where the local branch captures fine edges and the global branch maintains shape coherence.

Real3D-AD consists of single-sided scans with highly non-uniform point densities, where real defects frequently lie in self-occluded or sparsely sampled regions. Even under this challenging setting, PA3AD still achieves leading average performance and surpasses reconstruction-based methods on most categories. Additional experiments on more datasets are provided in Appendix A.

\begin{figure*}[!t]
	\centering
	\includegraphics[width=0.80\textwidth]{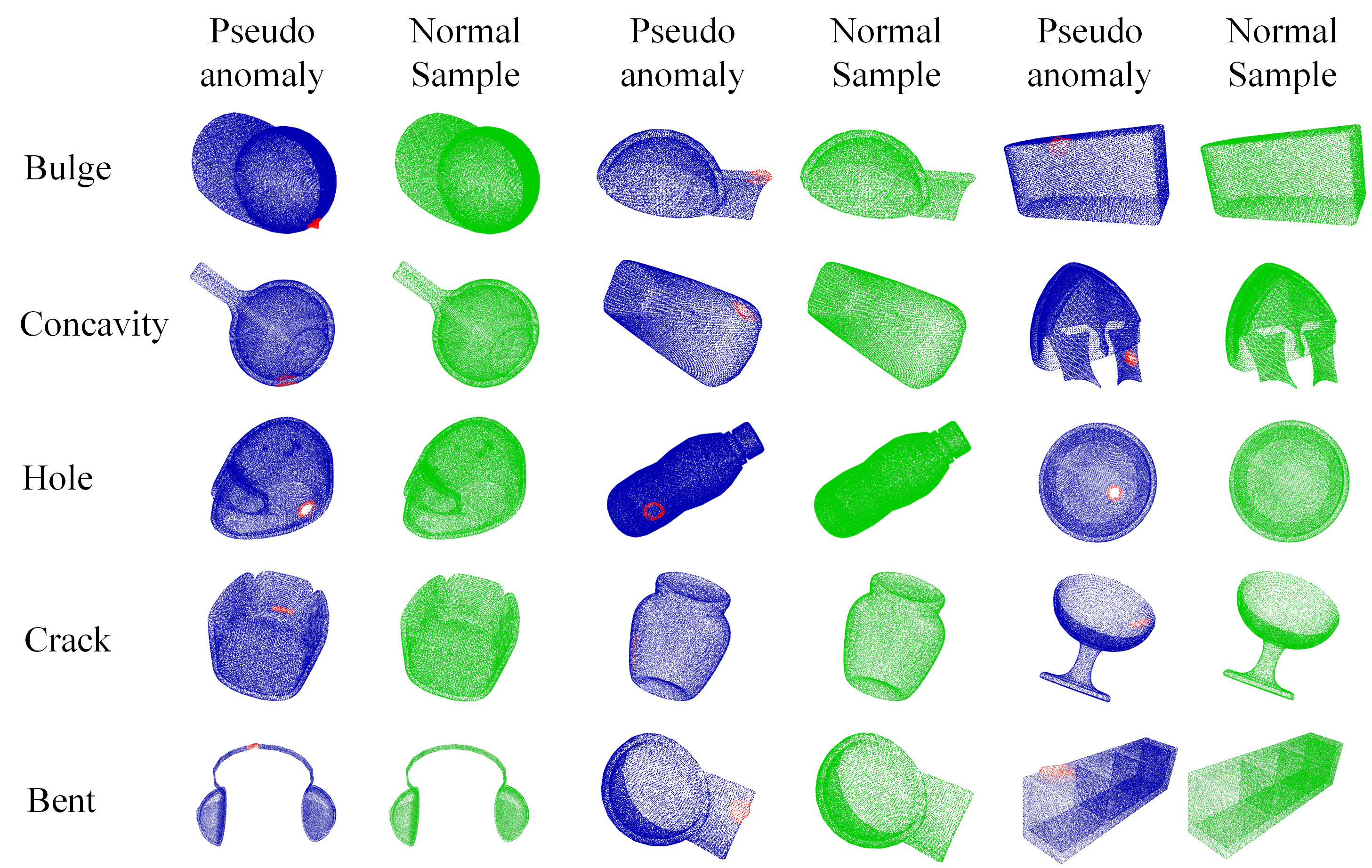}
	\caption{
		Visualization of pseudo‑anomaly generation. We show the anomalies generated using different physics-inspired deformation models and compare them with normal samples. Green represents normal samples, blue represents generated anomaly samples, and red represents abnormal areas.
	}
	\label{tab:pseudo‑anomaly generation}
\end{figure*}

\begin{figure*}[!t]
	\vspace{-10pt}
	\centering
	\includegraphics[width=0.88\linewidth]{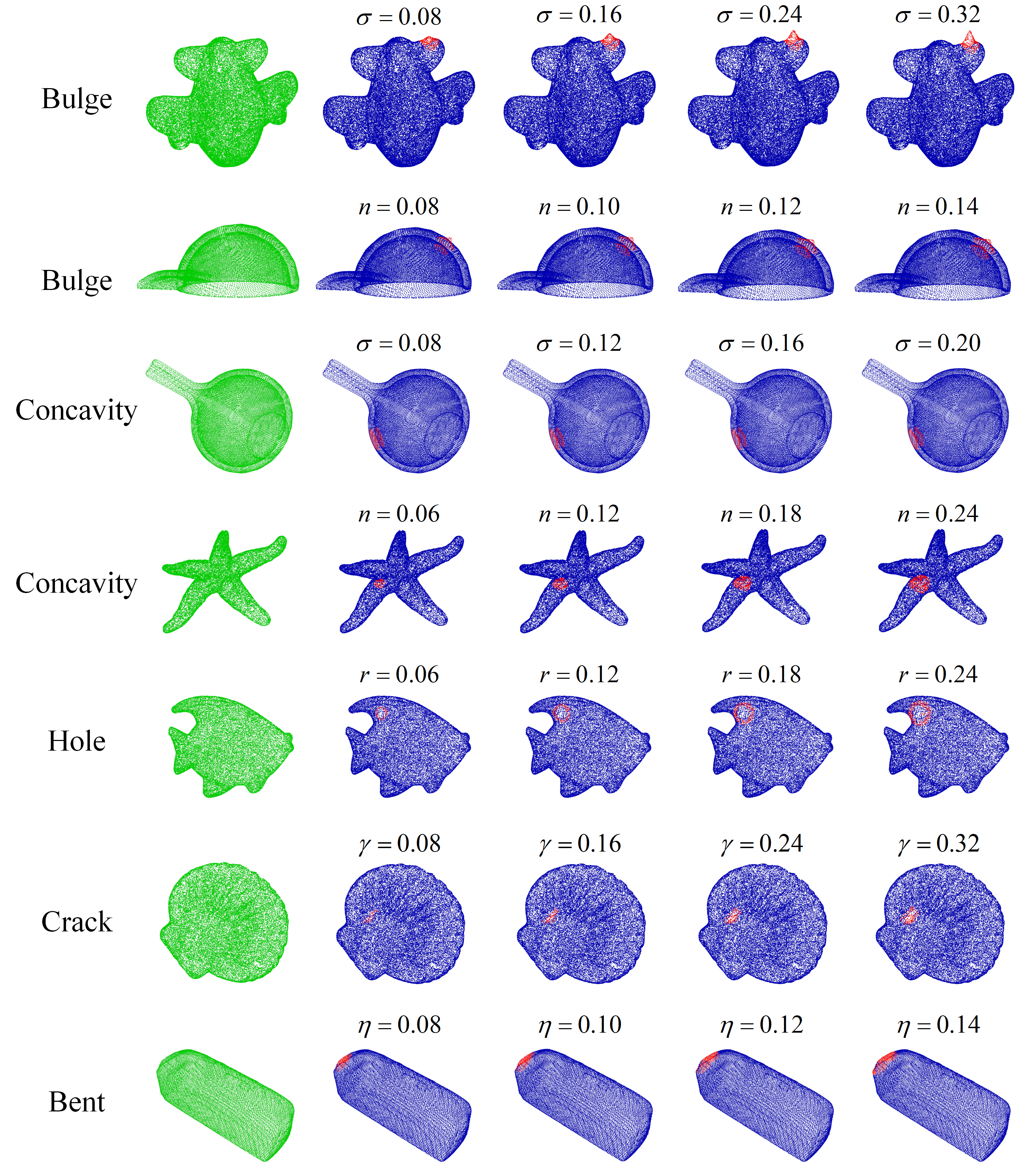}
	\caption{
		Visualization of parameter effects on pseudo-anomaly generation. The left column shows normal samples, and the right column shows pseudo-anomaly samples generated with different parameters. Different parameters control distinct anomaly morphologies (bulge, concavity, hole, crack, bent), achieving controllable pseudo-anomaly generation.
	}\label{vis}
\end{figure*}

\subsection{Pseudo-anomaly type ablations and generation visualization}
\label{section4.6}
We visualize the generated anomalies to assess their morphological plausibility and continuity. As shown in Fig.\ref{tab:pseudo‑anomaly generation}, our bulge-type and concavity-type methods perturb points along surface normals using a radial basis function, producing more natural continuous deformations. Hole-type and crack-type defects are modeled as approximately circular perforations and slit‑like through‑thickness openings. Bend-type is inspired by the Euler-Bernoulli deflection profile to reproduce smooth curvature changes of the point cloud.

%Fig.\ref{pseudo‑anomaly types ablation} reports an ablation on the choice and combination of pseudo‑anomaly types. We test AnomalyShapeNet by randomly selecting classes with IDs ending in 0. The results show that bulge and concavity contribute most to anomaly detection, likely due to differences in the dataset’s distribution of anomaly types. Overall, using all anomaly types yields the best performance. This validates the effectiveness of the method proposed in Section \ref{section3.3}. 

To demonstrate the impact of our MP-AG on pseudo-anomaly generation, we visualize the effect of each parameter on the generated anomalies. As shown in Fig.\ref{vis}, the left column shows the input normal point clouds, and the right columns show pseudo-anomaly samples generated with different hyperparameters. For bulge and concavity anomalies, parameters $\sigma$ and $n$ control the strength and spatial extent of the anomaly, respectively. Parameter $r$ controls the size of hole anomalies, parameter $\gamma$ controls the width of crack anomalies, and parameter $\eta$ controls the deformation degree of bent anomalies. Through these visualizations, we further demonstrate that controllable pseudo-anomaly generation yields anomalies that more closely align with physical intuition, resulting in improved anomaly detection performance.% More visualizations and experiments will be introduced in Appendix B and Appendix C.

\begin{figure*}[!t]
	\centering
	\includegraphics[width=1\textwidth]{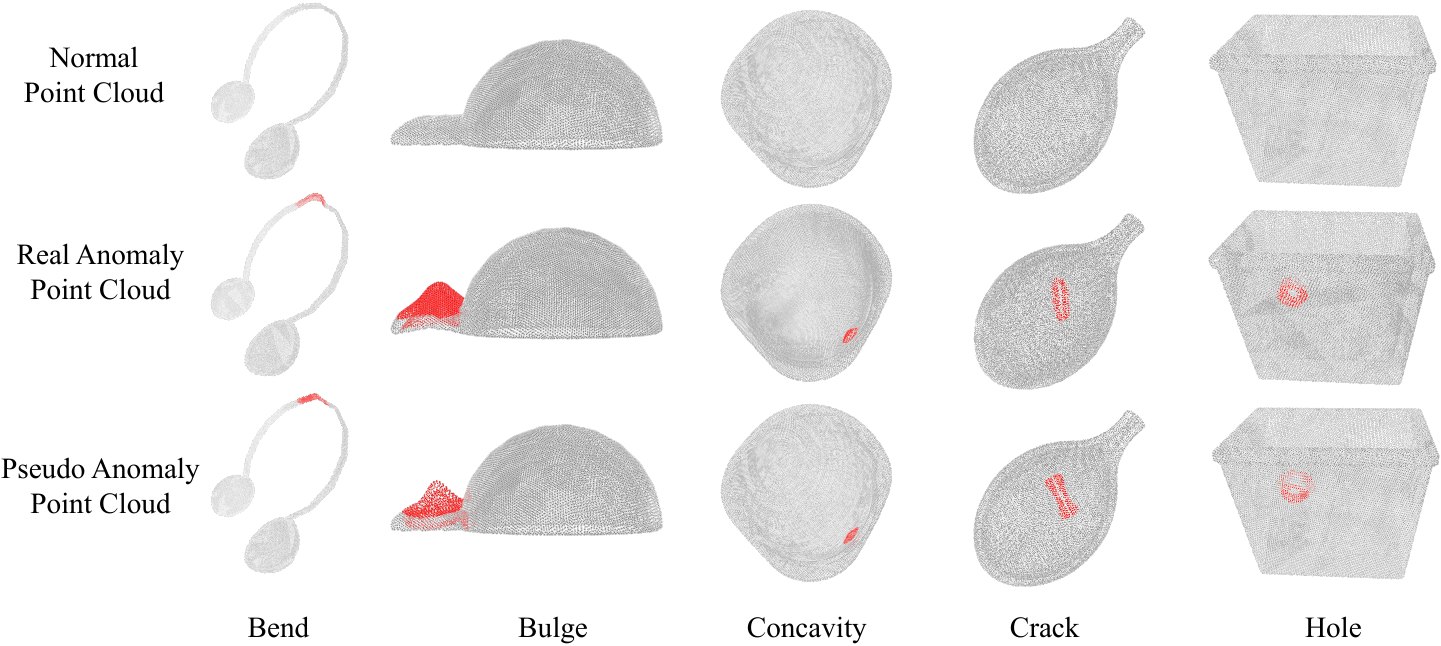}
	\caption{
		{Visual comparison between real anomalies and generated pseudo-anomalies. The three rows show normal point clouds, real anomaly point clouds, and pseudo-anomaly point clouds generated by MP-AG. 
			%The generated pseudo-anomalies exhibit morphologically similar deformation patterns and spatial distributions to their real counterparts, validating the effectiveness of our physics-inspired generation strategy.
			}
	}\label{fig:pseudo_vs_real}
\end{figure*}

{To further validate that our physics-inspired pseudo anomalies closely resemble real-world defects, we present a visual comparison in Fig.\ref{fig:pseudo_vs_real}. For five representative anomaly types (Bend, Bulge, Concavity, Crack, and Hole), we show normal point clouds, real anomaly point clouds from the dataset, and pseudo anomaly point clouds generated by MP-AG. The results demonstrate that our generated pseudo-anomalies exhibit morphologically similar deformation patterns and spatial distributions to their real counterparts, confirming that the physics-inspired design effectively captures the key geometric signatures of industrial defects.}

\subsection{Ablation on pseudo-anomaly generation strategy}
\label{section4.7}

To quantitatively validate the effectiveness of our physics-inspired generation strategy, we compare MP-AG with three perturbation baselines\cite{lian2024payload} under the same network architecture and training settings. As shown in Table \ref{tab:mpag_ablation}, MP-AG significantly outperforms all baselines on both datasets, demonstrating that preserving defect-specific morphological patterns provides higher-quality training supervision than unconstrained perturbations.

\begin{table}[h]
	\centering
	\small
	\renewcommand{\arraystretch}{0.80}
	\caption{Ablation on pseudo-anomaly generation strategies on Anomaly-ShapeNet and Real3D (O-AUROC).}
	\label{tab:mpag_ablation}
	\begin{tabular}{@{}lcccc@{}}
		\toprule
		\multirow{2}{*}{Generation Strategy} & \multicolumn{2}{c}{AnomalyShapeNet} & \multicolumn{2}{c}{Real3D} \\
		\cmidrule(lr){2-3} \cmidrule(lr){4-5}
		& bucket0 & vase0 & car & seahorse \\
		\midrule
		Random point deletion & 0.657 & 0.689 & 0.468 & 0.527 \\
		Random displacement & 0.711 & 0.718 & 0.402 & 0.473 \\
		Gaussian noise & 0.783 & 0.802 & 0.531 & 0.614 \\
		Ours & \textbf{0.968} & \textbf{0.975} & \textbf{0.769} & \textbf{0.875} \\
		\bottomrule
	\end{tabular}%
\end{table}

We further investigate the sensitivity of detection performance to the deformation magnitude. The default deformation parameters are uniformly sampled from $[0.08, 0.24]$. We uniformly scale all MP-AG parameters by a factor $k$ and report results in Table \ref{tab:param_sensitivity}. The default setting ($1.0\times$) achieves the best performance, while both under-scaled ($0.5\times$) and over-scaled ($2.0\times$, $5.0\times$) deformations lead to degraded detection.

\begin{table}[h]
	\centering
	\small
	\renewcommand{\arraystretch}{0.80}
	\caption{Sensitivity of detection performance to MP-AG deformation magnitude (O-AUROC).}
	\label{tab:param_sensitivity}
	\begin{tabular}{@{}lcccc@{}}
		\toprule
		\multirow{2}{*}{Scale factor} & \multicolumn{2}{c}{AnomalyShapeNet} & \multicolumn{2}{c}{Real3D} \\
		\cmidrule(lr){2-3} \cmidrule(lr){4-5}
		& headset0 & jar0 & fish & shell \\
		\midrule
		0.5$\times$ & 0.946 & 0.903 & 0.912 & 0.753 \\
		1.0$\times$ & \textbf{1.000} & \textbf{0.952} & \textbf{0.954} & \textbf{0.808} \\
		2.0$\times$ & 0.982 & 0.928 & 0.937 & 0.784 \\
		5.0$\times$ & 0.843 & 0.781 & 0.773 & 0.641 \\
		\bottomrule
	\end{tabular}%
\end{table}

\subsection{Experiments on prototype guidance and momentum-based updates}
\label{section4.8}

\begin{table}[h]
	\centering
	\small
	\renewcommand{\arraystretch}{0.8}
	\caption{Ablation on Prototype Guidance and Momentum Update on AnomalyShapeNet and Real3D.}
	\label{tab:pa3ad_wrap_left}
	\begin{tabular}{@{}ccccc@{}}
		\toprule
		\multirow{2}{*}{Method} & \multicolumn{2}{c}{AnomalyShapeNet} & \multicolumn{2}{c}{Real3D} \\
		\cmidrule(lr){2-3} \cmidrule(lr){4-5}
		& O-AUROC & P-AUROC & O-AUROC & P-AUROC \\
		\midrule
		$V_{1}$ & 0.874 & 0.865 & 0.724 & 0.772 \\
		$V_{2}$ & 0.921 & 0.915 & 0.768 & 0.815 \\
		$V_{3}$ & \textbf{0.936} & \textbf{0.924} & \textbf{0.789} & \textbf{0.837} \\
		\bottomrule
	\end{tabular}%
	
\end{table}

Table \ref{tab:pa3ad_wrap_left} evaluates the effects of prototype guided normal features and momentum based updates on anomaly localization and detection. $V_{1}$, $V_{2}$, and $V_{3}$ denote, respectively, training with pseudo‑anomalous samples only, guidance by prototype features, and guidance with momentum‑based updates. The experiments show that training with only pseudo‑anomalous samples leaves the model without a stable normal reference and prone to overfitting synthesis-specific features, resulting in reduced accuracy. Introducing prototype-guided normal features enhances feature discriminability. Moreover, momentum-based updating further stabilizes the prototype features, thereby improving anomaly-detection accuracy. These results validate the effectiveness of the methods proposed in Sections \ref{section3.4} and \ref{section3.5}.

\begin{table}[h]
	\centering
	\small
	\renewcommand{\arraystretch}{0.80}
	\caption{Sensitivity analysis of momentum scheduling strategies on Anomaly-ShapeNet (P-AUROC, mean$\pm$std over 5 runs).}
	\label{tab:momentum_ablation}
	\begin{tabular}{@{}lcc@{}}
		\toprule
		Momentum Strategy & ashtray0 & vase7 \\
		\midrule
		Fixed $\mu=0.1$ & 0.846$\pm$0.010 & 0.857$\pm$0.009 \\
		Fixed $\mu=0.999$ & 0.743$\pm$0.005 & 0.768$\pm$0.005 \\
		Cosine 0.1$\to$0.999 & 0.957$\pm$0.004 & 0.985$\pm$0.003 \\
		Linear 0.1$\to$0.999 (Ours) & \textbf{0.963$\pm$0.003} & \textbf{0.992$\pm$0.002} \\
		\bottomrule
	\end{tabular}%
\end{table}

We also analyze the sensitivity of the momentum scheduling strategy. Table \ref{tab:momentum_ablation} compares four strategies over 5 independent runs. A fixed low momentum ($\mu=0.1$) causes the prototype to fluctuate with each batch, leading to high variance. A fixed high momentum ($\mu=0.999$) causes the prototype to adapt too slowly in early training, resulting in the lowest performance. Both cosine and linear schedules achieve strong results by combining fast early adaptation with late-stage stability, with the linear schedule yielding the best mean and lowest variance.

\begin{table}[!htbp]
	\centering
	\small
	\renewcommand{\arraystretch}{0.80}
	\caption{Ablation of local and global attention on Anomaly-ShapeNet and Real3D.}
	\begin{tabular}{@{}lcccc@{}}
		\toprule
		\multirow{2}{*}{Method (I-AUROC)} & \multicolumn{2}{c}{Anomaly-ShapeNet} & \multicolumn{2}{c}{Real3D} \\
		\cmidrule(lr){2-3} \cmidrule(lr){4-5}
		& cap3 & vase3 & fish & seahorse \\
		\midrule
		baseline                  & 0.906 & 0.889 & 0.921 & 0.832 \\
		baseline+local            & 0.928 & 0.907 & 0.948 & 0.866 \\
		baseline+global           & 0.914 & 0.904 & 0.936 & 0.847 \\
		baseline+local+global     & \textbf{0.933} & \textbf{0.918} & \textbf{0.954} & \textbf{0.875} \\
		\bottomrule
	\end{tabular}
	\label{tab:local_global_wrap}
	
\end{table}

\subsection{Ablations on local-global attention}
\label{section4.9}

Table \ref{tab:local_global_wrap} presents experiments on the local-global attention mechanism, and we select two classes from each benchmark for evaluation.The baseline denotes no attention; local and global indicate the corresponding attention branches. The combined variant consistently yields the best AUROC. Local attention alone already accounts for most of the gains, indicating strong sensitivity to high-curvature rims, crack edges, and stamped textures. Global attention provides smaller but stable improvements, reflecting robustness to long-range structure, symmetry, and viewpoint changes.% More implementation details and additional ablation studies are presented in the Appendix D.

%\begin{wraptable}{r}{0.52\textwidth} 
%	\vspace{-8pt}
%	\small
%	\setlength{\tabcolsep}{6pt}
%	\begin{minipage}{\linewidth}
	%		\caption{Ablation of local and global attention on Anomaly-ShapeNet and Real3D.}
	%		\raggedright
	%		\begin{tabular}{@{}lcccc@{}}
		%			\toprule
		%			\multirow{2}{*}{Method (I-AUROC)} & \multicolumn{2}{c}{Anomaly-ShapeNet} & \multicolumn{2}{c}{Real3D} \\
		%			\cmidrule(lr){2-3} \cmidrule(lr){4-5}
		%			& cap3 & vase3 & fish & seahorse \\
		%			\midrule
		%			baseline                  & 0.906 & 0.889 & 0.921 & 0.832 \\
		%			baseline+local            & 0.928 & 0.907 & 0.948 & 0.866 \\
		%			baseline+global           & 0.914 & 0.904 & 0.936 & 0.847 \\
		%			baseline+local+global     & \textbf{0.933} & \textbf{0.918} & \textbf{0.954} & \textbf{0.875} \\
		%			\bottomrule
		%		\end{tabular}
	%		\label{tab:local_global_wrap}
	%	\end{minipage}
%	\vspace{-8pt}
%\end{wraptable}

\begin{table}[!htbp]
	\centering
	\small
	\renewcommand{\arraystretch}{0.80}
	\caption{Ablation of difference‑aware offset loss function (P-AUROC).}
	\begin{tabular}{@{}lcccc@{}}
		\toprule
		\multirow{2}{*}{Method (P-AUROC)} & \multicolumn{2}{c}{AnomalyShapeNet} & \multicolumn{2}{c}{Real3D} \\
		\cmidrule(lr){2-3} \cmidrule(lr){4-5}
		& bottle1 & vase1 & chicken & shell \\
		\midrule
		norm+dir                 & 0.861 & 0.873 & 0.824 & 0.787 \\
		norm+feat                & 0.601 & 0.552 & 0.484 & 0.496 \\
		dir+weight               & 0.542 & 0.561 & 0.471 & 0.503 \\
		norm+dir+weight          & 0.885 & 0.894 & 0.865 & 0.844 \\
		norm+dir+feat            & 0.908 & 0.929 & 0.882 & 0.869 \\
		norm+dir+feat+weight     & \textbf{0.915} & \textbf{0.939} & \textbf{0.890} & \textbf{0.880} \\
		\bottomrule
	\end{tabular}
	\label{tab:pa3ad_loss_wrap}
\end{table}

\subsection{Ablations on the loss function}
\label{section4.10}
%Table \ref{tab:pa3ad_loss_wrap} evaluates different loss combinations. The losses considered are offset‑norm, offset‑direction, feature‑discrepancy, and weight regularization. The experiments show that the offset‑norm and offset‑direction terms are foundational for learning accurate predictions, whereas the feature‑discrepancy and difference‑weight regularization terms enhance sample separability, leading to higher‑precision detection and localization. These results validate the effectiveness of the methods presented in Section \ref{section3.7}.

Table \ref{tab:pa3ad_loss_wrap} evaluates different loss combinations. The losses considered are offset‑norm, offset‑direction, feature‑discrepancy, and weight regularization. The experiments show that the offset‑norm and offset‑direction already form a strong baseline by aligning offset magnitude and orientation. Adding feature‑discrepancy consistently brings the largest improvement because it enlarges the gap between normal and anomalous features and raises recall on small defects. Weight regularization yields additional but smaller gains by preventing weight saturation and stabilizing learning under density and pose changes. Using all four losses gives the best accuracy, showing that accurate offset learning, explicit feature separation, and mild regularization are complementary and together improve detection and localization. These results validate the effectiveness of the methods presented in Section \ref{section3.7}.

\begin{figure*}[!ht]
	\centering
	\includegraphics[width=1.0\textwidth]{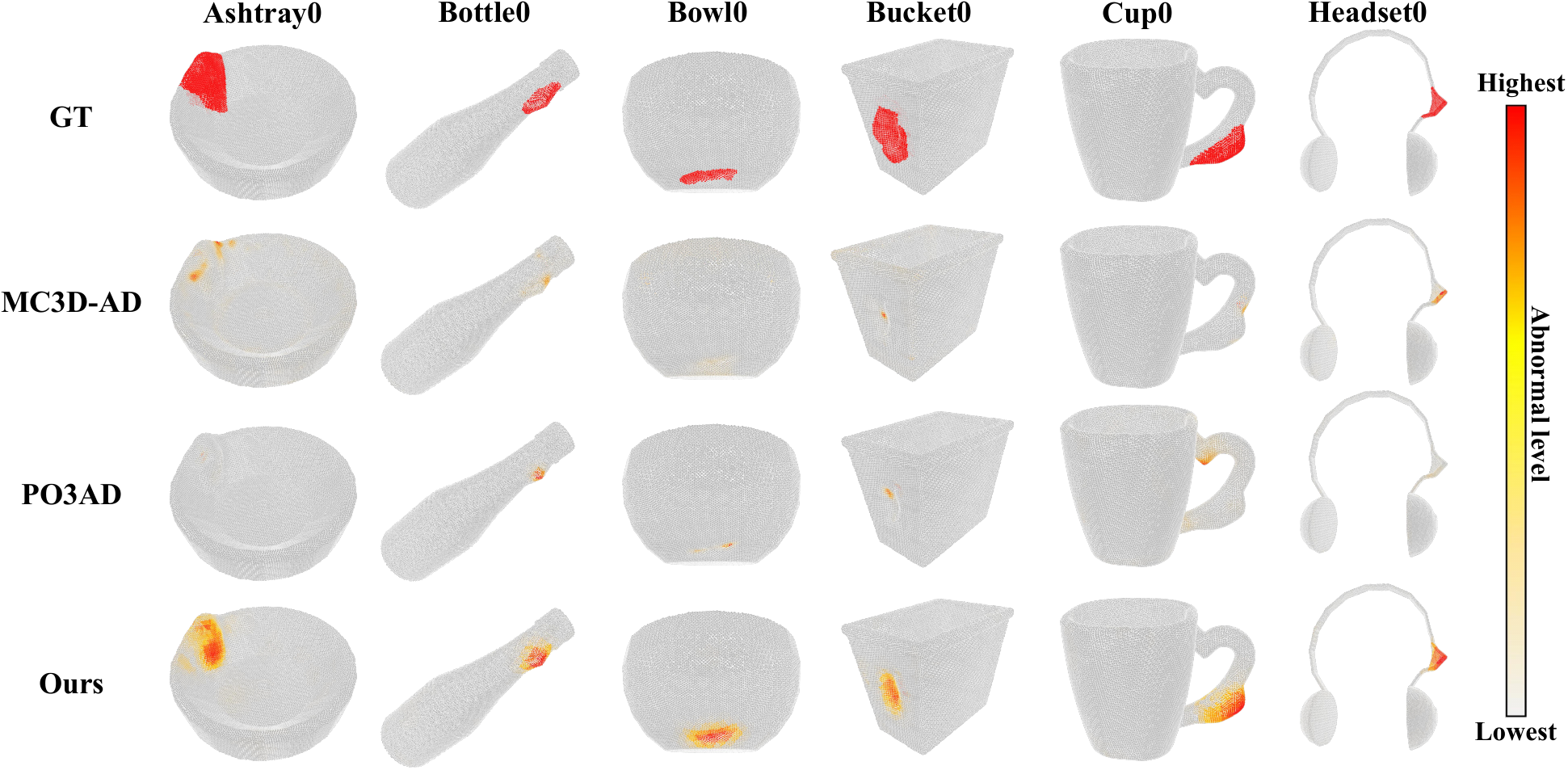}
	\caption{
	Qualitative comparison of anomaly localization on Anomaly-ShapeNet. We compare PA3AD with MC3D-AD and PO3AD on representative categories. For each method, point clouds are rendered with anomaly scores ranging from white to red, with the first row showing groundtruth annotations.
	}
	\label{anomaly localization}
\end{figure*}

\subsection{Visualization of prediction results on datasets}

As shown in Fig.\ref{anomaly localization}, we qualitatively compare PA3AD with MC3D-AD and PO3AD on representative categories from Anomaly-ShapeNet. For each method, point clouds are rendered with anomaly scores ranging from white (lowest) to red (highest). The visualizations demonstrate that PA3AD produces more precise localization with clear boundaries and fewer false positives in normal regions, even under complex geometry and sampling variations. These observations are consistent with the quantitative results and further validate the effectiveness of our approach.

\section{Conclusion}

%In this paper, we present a novel framework PA3AD for 3D point‑cloud anomaly detection based on prototype feature guidance and pseudo anomaly generation. We first synthesize diverse pseudo-anomalies by using multi-physics modeling. Next we use a shared‑weight network with local-global attention for stable feature extraction, and we employ momentum‑based prototype features to guide the learning of offsets between normal and anomalous samples. To further distinguish anomalous regions, we introduce a difference‑aware fusion block and a difference‑aware loss for accurate and robust object‑level and point‑level anomaly detection. Our PA3AD provides qualitative visualizations of the synthesized anomalies and localized defect regions, and attains state‑of‑the‑art average performance on the Anomaly‑ShapeNet and Real3D‑AD datasets.

In this paper, we proposed a novel framework, PA3AD, for 3D point-cloud anomaly detection based on prototype feature guidance and pseudo-anomaly generation. We designed a physics-inspired pseudo-anomaly generator that produces continuous and geometrically realistic defects, alleviating the scarcity of real anomalous samples and better matching real defect distributions. Normal and pseudo-anomalous point clouds were jointly encoded by a shared-weight backbone equipped with local-global attention, and prototype features were constructed with a momentum-based update scheme to provide a stable representation of normality. We further introduced a difference-aware fusion block to amplify feature discrepancies in anomalous regions and suppress responses in normal regions, and the resulting difference-enhanced features were mapped to pointwise geometric displacements and interpretable anomaly scores at both the object and point levels. PA3AD provides qualitative visualizations of synthesized anomalies and localized defect regions, and the results demonstrate state-of-the-art average performance on the Anomaly-ShapeNet and Real3D-AD datasets. In future work, we will further improve robustness under domain shifts and extend the framework to more diverse defect types and acquisition settings.

%\clearpage
%\newpage

\section*{CRediT authorship contribution statement}

\textbf{Jian Ning: } Writing - original draft, Conceptualization, Formal analysis, Methodology, Visualization; 
\textbf{Qin Zou: }Conceptualization, Funding acquisition, Project administration, Methodology, Resources, Supervision; 
\textbf{Linchun Wu: }Formal analysis, Methodology, Validation; 
\textbf{Yuanhao Yue: }Data curation, Formal analysis, Investigation, Validation; 
\textbf{Kunmo Li: }Conceptualization, Data curation, Visualization; 
\textbf{Shoubin Chen: }Formal analysis, Supervision, Validation, Writing – review \& editing; 
\textbf{Zhongyuan Wang: }Funding acquisition, Supervision, Writing – review \& editing.

\section*{Data availability}

Data will be made available on request.

\section*{Declaration of competing interest}

The authors declare that they have no known competing financial interests or personal relationships that could have appeared to influence the work reported in this paper.

\section*{Acknowledgements}

This research was supported by the Open Research Fund from Guangdong Laboratory of Artificial Intelligence and Digital Economy (SZ) under grant No. GML-KF-24-35, and National Natural Science Foundation of China under grant No. 62371350.

\newpage
\bibliographystyle{elsarticle-num}
\bibliography{natbib}

%% ========== Supplementary Material ==========
\newpage
\appendix
\setcounter{table}{0}
\setcounter{figure}{0}
\renewcommand{\thetable}{A\arabic{table}}
\renewcommand{\thefigure}{A\arabic{figure}}

\begin{center}
{\LARGE Supplementary Material \par}
\vspace{8pt}
\end{center}
\vspace{1em}

\section{Experiments on MVTec 3D-AD Dataset}

\begin{table*}[h!]
	\centering
	\renewcommand{\arraystretch}{0.80}
	\caption{Quantitative Results on MVTec 3D-AD Dataset. The Results are Presented in O-AUROC\%/P-AUROC\%. The best and second-place results are highlighted in \textcolor[rgb]{1, 0, 0}{\textbf{red}} and \textcolor[rgb]{0, .555, .976}{\textbf{blue}}.}
	\label{tab:mvtec3d}
	\begin{tabular}{l|ccc|c}
		\toprule
		Method $\rightarrow$ & BTF(Raw) & M3DM & CPMF & \textbf{PA3AD} \\
		Category $\downarrow$ & CVPRW'2023 & CVPR'2023 & PR'2024 & \textbf{Ours} \\
		\midrule
		Bagel       & 71.0/\textcolor[rgb]{1, 0, 0}{\textbf{97.3}} & 78.7/\textcolor[rgb]{0, .555, .976}{\textbf{96.2}} & \textcolor[rgb]{0, .555, .976}{\textbf{92.1}}/95.4 & \textcolor[rgb]{1, 0, 0}{\textbf{93.3}}/95.1 \\
		Cable Gland & 50.3/96.2 & 62.1/94.5 & \textcolor[rgb]{1, 0, 0}{\textbf{91.9}}/\textcolor[rgb]{0, .555, .976}{\textbf{97.7}} & \textcolor[rgb]{0, .555, .976}{\textbf{82.7}}/\textcolor[rgb]{1, 0, 0}{\textbf{98.3}} \\
		Carrot      & 78.3/\textcolor[rgb]{1, 0, 0}{\textbf{99.6}} & 60.5/98.6 & \textcolor[rgb]{1, 0, 0}{\textbf{99.0}}/\textcolor[rgb]{0, .555, .976}{\textbf{99.5}} & \textcolor[rgb]{0, .555, .976}{\textbf{93.7}}/97.6 \\
		Cookie      & 72.7/90.7 & \textcolor[rgb]{1, 0, 0}{\textbf{99.7}}/\textcolor[rgb]{0, .555, .976}{\textbf{93.0}} & 97.8/89.9 & \textcolor[rgb]{0, .555, .976}{\textbf{98.4}}/\textcolor[rgb]{1, 0, 0}{\textbf{94.5}} \\
		Dowel       & 90.8/94.9 & 79.5/94.3 & \textcolor[rgb]{0, .555, .976}{\textbf{92.8}}/\textcolor[rgb]{0, .555, .976}{\textbf{95.1}} & \textcolor[rgb]{1, 0, 0}{\textbf{96.7}}/\textcolor[rgb]{1, 0, 0}{\textbf{95.2}} \\
		Foam        & 53.2/\textcolor[rgb]{0, .555, .976}{\textbf{93.3}} & 74.9/92.5 & \textcolor[rgb]{0, .555, .976}{\textbf{76.4}}/\textcolor[rgb]{1, 0, 0}{\textbf{94.2}} & \textcolor[rgb]{1, 0, 0}{\textbf{87.3}}/93.1 \\
		Peach       & 57.6/\textcolor[rgb]{1, 0, 0}{\textbf{98.7}} & 66.6/95.5 & \textcolor[rgb]{0, .555, .976}{\textbf{92.3}}/\textcolor[rgb]{1, 0, 0}{\textbf{98.7}} & \textcolor[rgb]{1, 0, 0}{\textbf{92.6}}/\textcolor[rgb]{0, .555, .976}{\textbf{97.8}} \\
		Potato      & 64.2/\textcolor[rgb]{1, 0, 0}{\textbf{99.9}} & 49.6/98.2 & \textcolor[rgb]{1, 0, 0}{\textbf{98.2}}/\textcolor[rgb]{0, .555, .976}{\textbf{99.7}} & \textcolor[rgb]{0, .555, .976}{\textbf{97.8}}/98.9 \\
		Rope        & 93.2/\textcolor[rgb]{0, .555, .976}{\textbf{98.9}} & 91.6/\textcolor[rgb]{1, 0, 0}{\textbf{99.1}} & \textcolor[rgb]{0, .555, .976}{\textbf{94.9}}/98.7 & \textcolor[rgb]{1, 0, 0}{\textbf{96.9}}/98.6 \\
		Tire        & 50.2/\textcolor[rgb]{0, .555, .976}{\textbf{98.8}} & 60.5/98.5 & 82.2/\textcolor[rgb]{1, 0, 0}{\textbf{98.9}} & \textcolor[rgb]{1, 0, 0}{\textbf{94.1}}/96.3 \\
		\midrule
		Mean        & 68.2/\textcolor[rgb]{1, 0, 0}{\textbf{96.8}} & 72.4/96.0 & \textcolor[rgb]{0, .555, .976}{\textbf{91.8}}/\textcolor[rgb]{1, 0, 0}{\textbf{96.8}} & \textcolor[rgb]{1, 0, 0}{\textbf{93.3}}/\textcolor[rgb]{0, .555, .976}{\textbf{96.5}} \\
		\bottomrule
	\end{tabular}
\end{table*}

The MVTec 3D-AD dataset is a comprehensive benchmark for 3D anomaly detection in industrial scenarios. It contains 10 object categories (Bagel, Cable Gland, Carrot, Cookie, Dowel, Foam, Peach, Potato, Rope, and Tire), each consisting of high-resolution 3D point cloud scans acquired under controlled conditions. The dataset provides both normal training samples and anomalous test samples with pixel-level ground truth annotations, covering a variety of real-world surface and structural defects. It has become one of the most widely used benchmarks for evaluating unsupervised 3D anomaly detection methods.

To evaluate the generalization ability of our method on this dataset, we compare PA3AD with three representative methods: BTF, M3DM, and CPMF. We report the object-level AUROC (O-AUROC) and point-level AUROC (P-AUROC) for each category. The quantitative results are presented in Table~\ref{tab:mvtec3d}.

As shown in Table~\ref{tab:mvtec3d}, our PA3AD achieves the highest mean O-AUROC of 93.3\%, outperforming the second-best method CPMF by 1.5\%. For O-AUROC, PA3AD ranks first in 6 out of 10 categories (Bagel, Dowel, Foam, Peach, Rope, and Tire), demonstrating strong object-level detection capability. For P-AUROC, although the overall mean is competitive (96.5\%), the point-level localization on this dataset remains challenging due to the relatively small anomaly regions in certain categories. These results confirm that PA3AD generalizes well to the MVTec 3D-AD benchmark beyond the primary datasets used in our main experiments.

\section{Pseudo-anomaly generation visualization}
\subsection{Visualization on the Real3D-AD dataset}

\begin{figure*}[!t]
	\centering
	\includegraphics[width=0.80\textwidth]{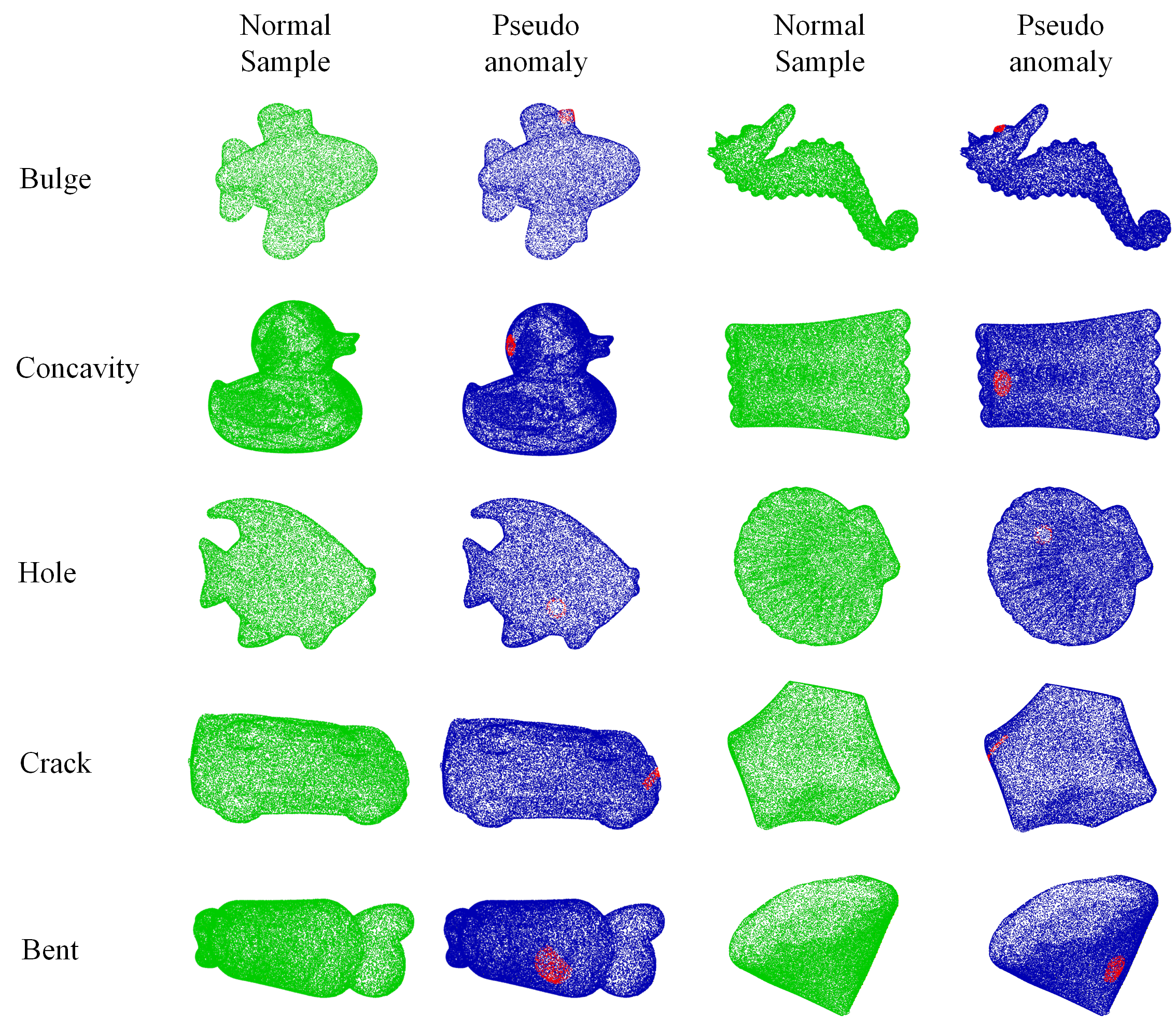}
	\caption{
		Visualization of pseudo-anomaly generation on the Real3D-AD dataset. We show the anomalies generated using different physical models and compare them with normal samples. Green represents normal samples, blue represents generated anomaly samples, and red represents abnormal areas.  
	}\label{Real3d}
\end{figure*}

As shown in Fig.~\ref{Real3d}, we visualize the pseudo-anomaly generation results on the Real3D-AD dataset to qualitatively assess the realism and diversity of the synthesized defects. The generated anomalies cover a wide range of geometric patterns, including bulge, concavity, hole, crack, and bent, which are representative of real-world surface and structural degradations in industrial scenarios. In addition to the diversity of anomaly categories, the synthesized defects exhibit continuous deformation characteristics rather than isolated, noisy artifacts, indicating that the generation process preserves plausible geometric transitions. Moreover, the anomalies are spatially coherent with the underlying object structure: their shapes, orientations, and extents adapt naturally to local surfaces and edges, and the surrounding regions remain visually consistent. Overall, these visualizations suggest that our method is able to produce realistic, high-quality pseudo anomalies that closely resemble real defects, thereby providing informative and reliable samples for subsequent anomaly detection and localization.

\begin{figure}[!h]
	\centering
	\includegraphics[width=0.85\linewidth]{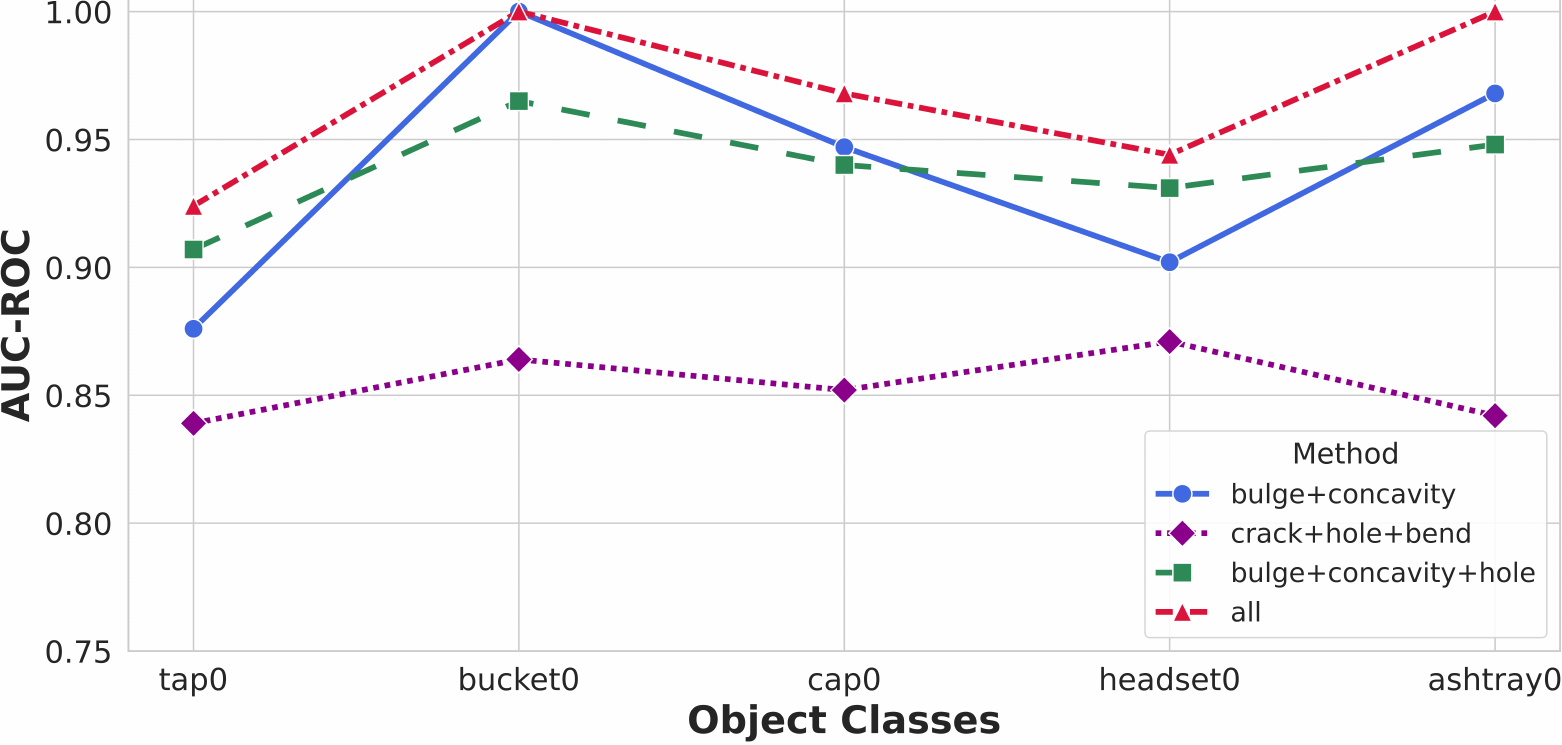}
	\caption{Ablation experiments of different pseudo-anomaly types.}
	\label{pseudo-anomaly types ablation}
\end{figure}

\subsection{Ablation experiments of pseudo-anomaly types}

Fig.~\ref{pseudo-anomaly types ablation} presents an ablation study on the selection and combination of pseudo anomaly types used during training. To ensure a fair and reproducible evaluation, we test AnomalyShapeNet on a held-out subset formed by randomly selecting categories whose class IDs end with 0, while keeping all other settings unchanged. We compare using a single pseudo-anomaly type versus combining multiple types, and report the corresponding detection performance. The results indicate that the bulge and concavity types provide the most consistent gains, suggesting that these deformations better match the dominant anomaly patterns (or are more distinguishable from normal geometry) under the distribution of the AnomalyShapeNet dataset. In contrast, other types such as hole, crack, and bent tend to yield smaller improvements when used alone, possibly because they appear less frequently, are more subtle at the point level, or are harder to synthesize with stable geometric cues. Importantly, when all pseudo-anomaly types are jointly employed, the model achieves the best overall performance. This confirms that different anomaly types offer complementary supervision signals, improving both the generalization to diverse defect patterns and the robustness of the detector across categories.

\section{Precision-recall analysis}

\begin{figure}[!h]
	\centering
	\includegraphics[width=\columnwidth]{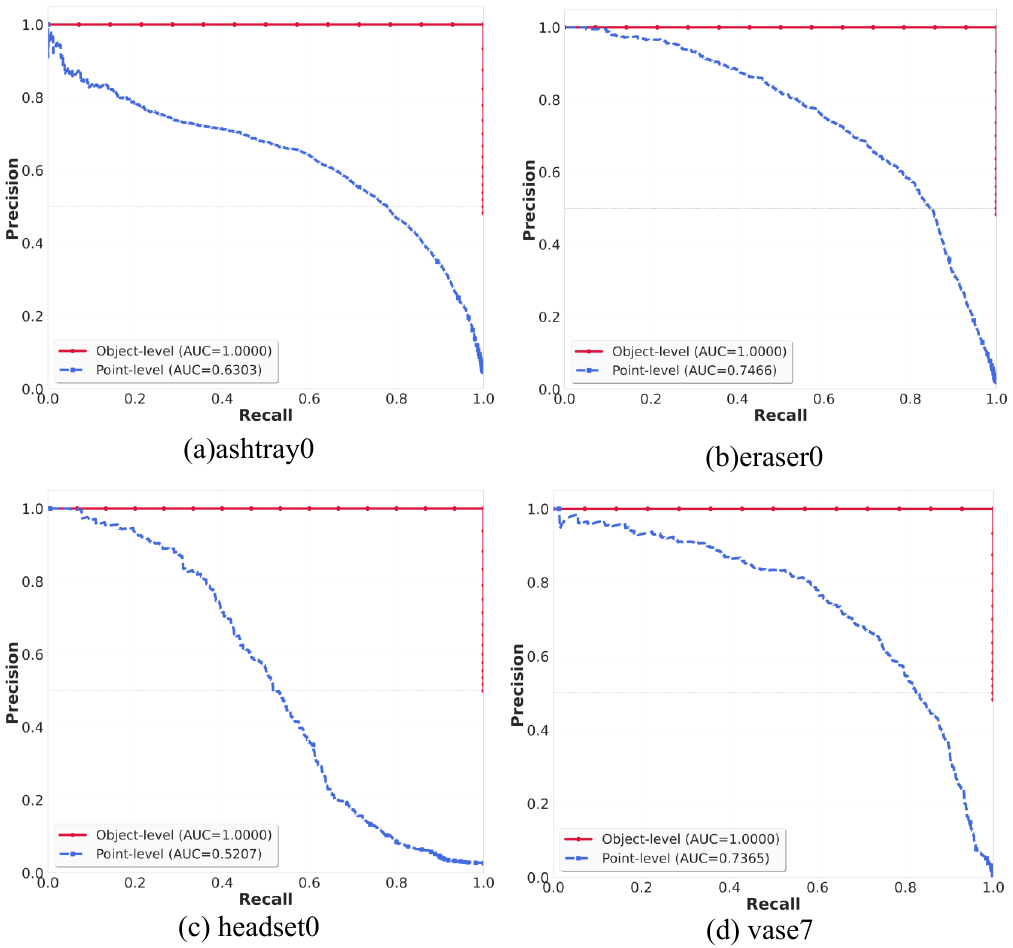}
	\caption{
		Precision-recall curves on AnomalyShapeNet. 
	}\label{Precision-recall}
\end{figure}

To further analyze the detection performance of our model, we plot the precision-recall curves for four representative categories (ashtray0, eraser0, headset0, vase7) on the AnomalyShapeNet dataset, as shown in Fig.\ref{Precision-recall}. Our method achieves almost perfect object-level results, with the object-level AUC-PR reaching 1.00. In contrast, the point-level PR curves gradually decline as recall increases, indicating the higher difficulty of fine-grained point-wise anomaly localization, while still maintaining satisfactory AUC-PR performance overall. These results show that our method can not only reliably distinguish normal and anomalous objects, but also provide meaningful localization of anomalous regions.

\section{Ablation Study on Loss Functions}

\begin{table}[h]
	\centering
	\small
	\caption{Ablation of loss function hyperparameters on AnomalyShapeNet and Real3D dataset.}
	\label{tab:pa3ad_loss_hyper}
	\resizebox{\columnwidth}{!}{%
	\begin{tabular}{@{}cccccccc@{}}
		\toprule
		\multirow{2}{*}{norm} & \multirow{2}{*}{dir} & \multirow{2}{*}{feat} & \multirow{2}{*}{weight} & \multicolumn{2}{c}{AnomalyShapeNet} & \multicolumn{2}{c}{Real3D} \\
		\cmidrule(lr){5-6} \cmidrule(lr){7-8}
		& & & & O-AUROC & P-AUROC & O-AUROC & P-AUROC \\
		\midrule
		\checkmark & \checkmark & 0.10  & 0.001 & 0.917 & 0.907 & 0.775 & 0.827 \\
		\checkmark & \checkmark & 0.01  & 0.10  & 0.902 & 0.893 & 0.762 & 0.806 \\
		\checkmark & \checkmark & 0.01  & 0.01  & 0.914 & 0.903 & 0.777 & 0.822 \\
		\checkmark & \checkmark & \textbf{0.01}  & \textbf{0.001} & \textbf{0.936} & \textbf{0.924} & \textbf{0.789} & \textbf{0.837} \\
		\checkmark & \checkmark & 0.001 & 0.001 & 0.923 & 0.917 & 0.784 & 0.833 \\
		\bottomrule
	\end{tabular}%
	}
\end{table}

In the ablation study on loss function hyperparameters, we compare different combinations of $\lambda_{\text{feat}}$ and $\lambda_{\text{weight}}$ on the AnomalyShapeNet and Real3D datasets, as shown in Table~\ref{tab:pa3ad_loss_hyper}, where \text{norm} and \text{dir} are always enabled. The results show that when $\lambda_{\text{feat}} = 0.01$ and $\lambda_{\text{weight}} = 0.001$, our method achieves the highest O-AUROC and P-AUROC on both datasets. In contrast, further increasing $\lambda_{\text{weight}}$ leads to a clear performance degradation, while decreasing $\lambda_{\text{feat}}$ does not bring additional gains. Overall, a moderate feature-term weight and a smaller weighting coefficient better balance the contributions of different loss components, resulting in more stable and robust detection and localization performance.

\section{Computational Cost Comparison}

\begin{table}[h]
	\centering
	\small
	\setlength{\tabcolsep}{3pt}
	\caption{{Computational cost comparison with representative baselines. Infer Time and GPU Mem are measured on a single NVIDIA RTX 4090 GPU.}}
	\label{tab:computational_cost}
	\resizebox{\columnwidth}{!}{%
	\begin{tabular}{@{}lccccc@{}}
		\toprule
		Method & Input Points & Params (M) & Trainable (M) & Infer Time (ms) & GPU Mem (GB) \\
		\midrule
		r3d-ad & 2048 & 1.58 & 1.58 & 72.96 & 0.03 \\
		Reg2Inv & 8192 & 5.24 & 5.24 & 341.77 & 5.14 \\
		PO3AD & $\sim$44K & 37.86 & 37.86 & 12.94 & 0.17 \\
		PASDF & $\sim$44K & 0.40 & 0.40 & 3.05 & 0.09 \\
		MC3D-AD & $\sim$44K & 31.03 & 8.93 & 58.55 & 0.88 \\
		PA3AD (Ours) & $\sim$44K & 78.88 & 78.88 & 13.13 & 0.36 \\
		\bottomrule
	\end{tabular}%
	}
\end{table}

{Table~\ref{tab:computational_cost} compares the computational cost of PA3AD with representative baselines. Due to differences in network architecture, some methods cannot process full-resolution point clouds ($\sim$44K points) as input: r3d-ad requires downsampling to 2048 points, and Reg2Inv operates on paired point subsets of 8192 points. Although PA3AD has a larger parameter count due to its backbone architecture, its efficient design enables fast inference (13.13 ms) with low GPU memory consumption (0.36 GB), demonstrating competitive practical efficiency among methods that support full-resolution input.}

\end{document}